\documentclass[10pt,twocolumn,letterpaper]{article}

\usepackage{cvpr}              
\usepackage{multirow}
\usepackage[table]{xcolor}
\usepackage{pifont}

\definecolor{cvprblue}{rgb}{0.21,0.49,0.74}
\definecolor{BrickRedTitle}{rgb}{0.70, 0.13, 0.13}
\usepackage[pagebackref,breaklinks,colorlinks,allcolors=cvprblue]{hyperref}

\def\paperID{5088} 
\def\confName{CVPR}
\def\confYear{2026}

\title{HERO: Hierarchical Traversable 3D Scene Graphs for Embodied Navigation Among Movable Obstacles}

\author{
  Yunheng Wang$^{1,*}$
  \space\space
  Yixiao Feng$^{1,*}$
  \space\space
  Yuetong Fang$^{1,*}$
  \space\space
  Shuning Zhang$^{1}$
  \space\space
  Tan Jing$^{1}$
  \space\space
  Jian Li$^{1}$
  \\
  Xiangrui Jiang$^{1}$
  \space\space
  Renjing Xu$^{1,\dagger}$
  \\[2pt]
  \small{\textnormal{$^{1}$The Hong Kong University of Science and Technology (Guangzhou)}}\\
    \small{$^{*}$Equal contribution\space\space $^{\dagger}$Corresponding author} \\
}

\begin{document}

\twocolumn[{
\maketitle
\begin{center}
    \includegraphics[width=\linewidth]{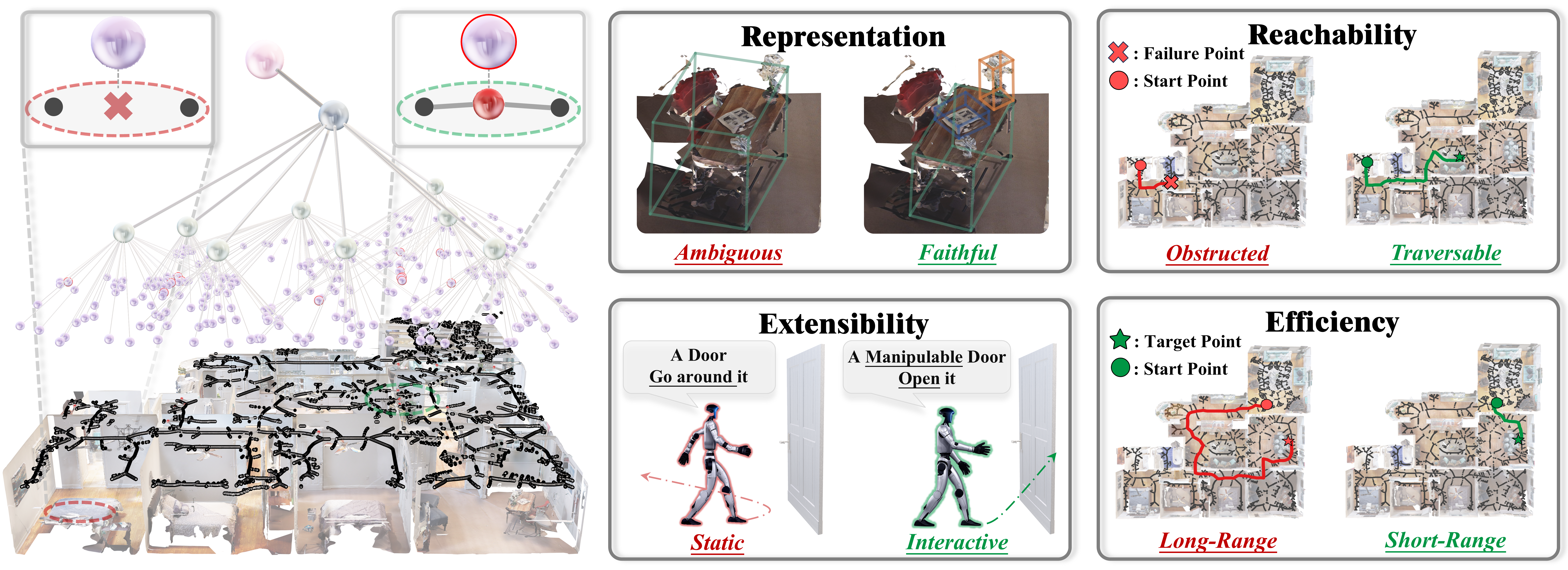}
\end{center}
\vspace{-0.5cm}
\captionsetup{type=figure}
\captionof{figure}{
    HERO builds Hierarchical Traversable 3D Scene Graphs that capture scene structure, object semantics, and functional movability, thereby enabling more faithful \textbf{representation} and interactive \textbf{extensibility} in complex physical environments. By explicitly encoding object interactivity within the navigation graph, HERO incorporates movable obstacles and redefines the traversable space, ultimately achieving higher \textbf{reachability} and more \textbf{efficient} navigation behaviors.
}
\label{fig:teaser}
\vspace{0.5 cm}
}]

\begin{center}
  \large\bfseries Abstract
\end{center}

\textit{3D Scene Graphs (3DSGs) constitute a powerful representation of the physical world, distinguished by their abilities to explicitly model the complex spatial, semantic, and functional relationships between entities, rendering a foundational understanding that enables agents to interact intelligently with their environment and execute versatile behaviors. Embodied navigation, as a crucial component of such capabilities, leverages the compact and expressive nature of 3DSGs to enable long-horizon reasoning and planning in complex, large-scale environments. However, prior works rely on a static-world assumption, defining traversable space solely based on static spatial layouts and thereby treating interactable obstacles as non-traversable. This fundamental limitation severely undermines their effectiveness in real-world scenarios, leading to limited reachability, low efficiency, and inferior extensibility. To address these issues, we propose HERO, a novel framework for constructing Hierarchical Traversable 3DSGs, that redefines traversability by modeling operable obstacles as pathways, capturing their physical interactivity, functional semantics, and the scene's relational hierarchy. The results show that, relative to its baseline, HERO reduces PL by \textbf{35.1\%} in partially obstructed environments and increases SR by \textbf{79.4\%} in fully obstructed ones, demonstrating substantially higher efficiency and reachability.}

\section{Introduction}

Autonomous robots executing high-level tasks require scene understanding that transcend the purely geometric maps from conventional 3D reconstruction~\cite{SLAM, SFM}. 3D Scene Graphs (3DSGs) address this gap by providing a powerful abstraction that explicitly models semantic constituents in a scene and their structured spatial-topological constraints, enabling human-aligned reasoning~\cite{3DGIR, ASG3DSG}. 
While early flat 3DSGs focused on local object-to-object relations~\cite{SceneGraphFusion, Conceptgraphs}, Hierarchical 3D Scene Graphs (H-3DSGs) represent a significant advancement. The core advantage of H-3DSGs is their organization of environments across multiple spatial-semantic levels (e.g., objects to rooms to floors). This hierarchical structure is crucial for embodied navigation, as it supports the coherent reasoning and long-range planning required for composite tasks~\cite{FSRVLN, HOVSG, OpenGraph, Point2Graph}. 

However, despite their hierarchical advantages, most existing H‑3DSG approaches~\cite{HOVSG, HiDynaGraph, OVIOGW3DHSG, FSRVLN} still share a critical limitation. They are built upon an open‑world assumption, namely that the environment is fully accessible from the outset and that the navigation graph can be constructed as if all regions marked as traversable by the current scene layout were already open and unobstructed. In real life, this assumption is often difficult to hold true. Various obstacles frequently exist in the environment, such as doors, curtains, and movable barriers, all of which can obstruct the entire passageway. Traditional methods simplify the properties of objects in a scene, treating them like static walls, thus ignoring their interactivity or manipulability and incorrectly defining these obstacles as permanently insurmountable.

This rigid interpretation of obstacles leads to significant limitations:
~\ding{172} Inferior representation and extensibility: The functional‐attribute homogenization of objects fundamentally restricts the robot’s capability to accomplish intelligent and compositional tasks in complex environments;
~\ding{174} Limited reachability: The presence of obstructing objects constrains the robot’s accessible space, making certain target regions physically unreachable despite being spatially proximate;
~\ding{173} Low efficiency: Obstacle-avoidance planning that strictly enforces collision-free constraints yields overly conservative free-space estimation, inducing excessive detours and suboptimal trajectories.

Motivated by these limitations, we revisit obstacle representation in H‑3DSGs through the lens of human navigation. Rather than uniformly modeling all obstacles as rigid and impassable, we draw inspiration from how humans perceive and navigate their surroundings \cite{HumanNav}. In real environments, humans do not regard all blocking objects as absolute barriers; instead, they instinctively evaluate the object's properties and potential affordances. While immovable structures necessitate detours, objects such as lightweight items, movable furniture, or operable doors can be manipulated to enable direct passage. This natural ability supports more flexible, efficient, and goal-driven navigation.

In this work, we present \textbf{HERO}, a \textbf{H}ierarchical Trav\textbf{er}sable 3DSG for emb\textbf{o}died navigation. The contributions are summarized as follows:

\begin{enumerate}
\item A three-stage framework that jointly extracts geometric structure, semantic attributes, and physical interactivity, and integrates them into a unified H-3DSG with substantially enriched representational capacity.

\item Three dedicated strategies that enhance the accuracy of semantic and interactivity representations, effectively mitigating cross-room visual interference and strengthening the consistency of object-level semantics.

\item An obstacle-aware navigation formulation that incorporates physically movable obstacles into the navigation graph and selects candidates by their contribution to path optimality, thereby redefining traversable regions and improving navigation reachability and efficiency.

\end{enumerate}
\begin{figure*}[t]
    \centering
    \includegraphics[width=\linewidth]{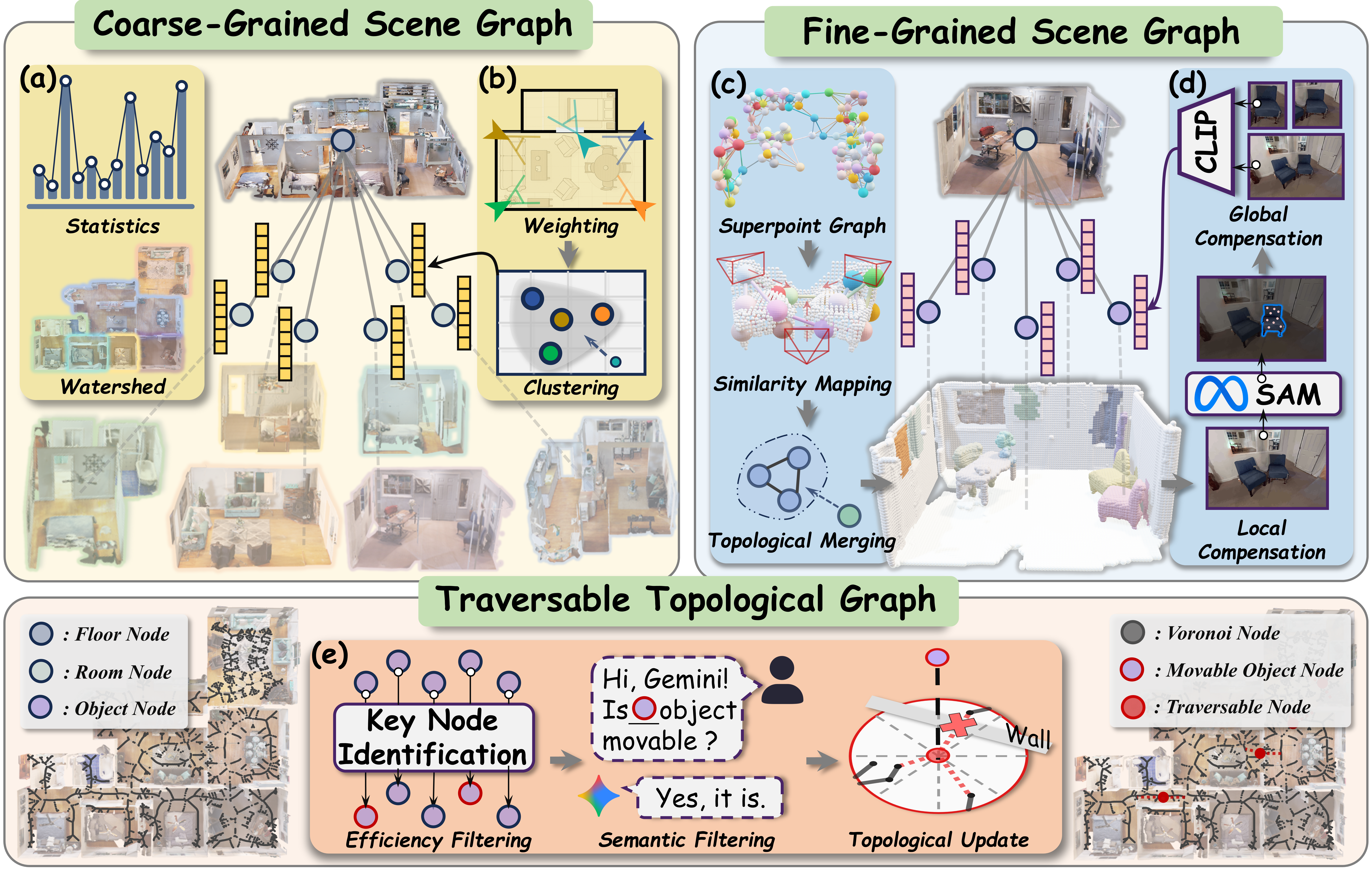}
    \vspace{-0.5cm}
    \caption{
        {\textbf{Pipeline of HERO.} We construct the 3D scene graph in three stages. 
        The Coarse-Grained Scene Graph is derived through \textbf{(a)} geometric decomposition to obtain floor–room structures, followed by the \textbf{(b)} Visibility Purification Strategy to produce room-consistent semantic representations.
        The Fine-Grained Scene Graph is obtained using the \textbf{(c-d)} Topological Clustering Strategy, which forms geometrically coherent object nodes and refines their semantic attributes.
        The Traversable Topological Graph is finally constructed by the \textbf{(e)} Traversability Update Strategy, modeling interactive traversability by integrating movable obstacles}
    }
    \label{fig:main_figure}
    \vspace{-0.2cm}
\end{figure*}



\section{Related Work}

\subsection{3D Scene Graphs}
Early efforts on 3DSGs starting from~\cite{3DSGs} introduced the idea of representing complex environments through a graph that jointly encodes geometric structure, object-level semantics, and inter-object spatial relationships. Such representations provide robots with a structured understanding that supports spatial reasoning, multi-step planning, and long-horizon navigation. However, these early 3DSGs~\cite{SceneGraphFusion, 3DSGs} rely on closed-set semantics, limiting their ability to generalize to previously unseen categories and constraining their utility in open-world robotic applications. This limitation motivated a series of open-vocabulary 3DSG research~\cite{ConceptFusion, Conceptgraphs, HOVSG, OpenGraph, Point2Graph, chen2024clip, koch2024open3dsg}. Among them, ConceptFusion~\cite{ConceptFusion} and ConceptGraphs~\cite{Conceptgraphs} focus primarily on object-level or instance-level scene graphs, achieving open-vocabulary labeling but lacking higher-level abstraction such as rooms, floors, and functional regions, that restricts efficient object retrieval and hinders large-scale navigation. 

To address these shortcomings, recent works proposed H-3DSGs, explicitly incorporating multi-scale semantics and extending applicability to both indoor~\cite{HOVSG, FSRVLN} and outdoor environments~\cite{OpenGraph, IAO3DSGVLESO}. HOV-SG~\cite{HOVSG} provides a representative formulation by constructing a floor–room–object hierarchy enriched with open-vocabulary semantics, enabling efficient object retrieval and long-horizon language-guided navigation in multi-story indoor environments. Building on this foundation, H-3DSGs have since been extended to a wide range of applications, including autonomous parking~\cite{Parking-SG}, multi-agent collaboration~\cite{tan2025roboos, chang2024partnr}, and embodied mobile manipulation~\cite{honerkamp2024language, yan2025dynamic}, indicating the growing importance and generality of hierarchical scene abstractions across real-world robotic systems.

\subsection{Navigation Among Movable Obstacles} 
Navigation Among Movable Obstacles (NAMO)~\cite{NAMO} endows robots with the ability to actively reshape their surroundings, forming a crucial competency for complex, long-horizon tasks. A core challenge lies in accurately inferring object traversability and integrating this reasoning into global navigation decisions. Early NAMO approaches relied heavily on geometric search, hand-crafted priors~\cite{NAMOWOLUPS, JPAPPAMO, NAMOUNC}, or rule-based assumptions~\cite{ENAMOUAMMVHPL, PTCWMAOBO, IMAGE2MASS, SWIPEBOT} to characterize obstacle movability, which limited their robustness and generalization beyond simplified settings. To overcome these constraints, subsequent research incorporated richer perceptual cues, such as learned movability prediction~\cite{IFAR, AINOQRULLM}, tactile feedback~\cite{TBNOUODNIUEWMO}, and affordance estimation~\cite{ABMRNAMO}, enabling robots to autonomously infer object traversability through interactive perception. However, despite improving robustness in unstructured environments, interactive perception intrinsically requires physical contact, introducing risk, slowing down exploration, and making it difficult to seamlessly incorporate real-time traversability judgments into high-level planning.
Recently, several studies have demonstrated the feasibility of non-contact paradigms that leverage the reasoning capabilities of foundation models~\cite{INIRWTOULLAVLM, NAMO_LLM}. Nevertheless, these methods still depend on exhaustive object pre-identification and typically operate outside the global planning loop, limiting their applicability to long-horizon NAMO decision-making. To address these challenges, our approach constructs a Hierarchical Traversable 3DSG that serves as a unified substrate for high-level planning. This representation embeds actionable traversability cues directly into a multi-level scene structure, enabling efficient long-horizon decision-making in interaction-rich environments while drastically reducing dependence on explicit object pre-identification.

\section{Method}
We formulate the Hierarchical Traversable 3D Scene Graph representation (Sec.~\ref{sec:Problem_Formulation_Overview}) and present HERO, a framework for its systematic construction. As illustrated in Fig.~\ref{fig:main_figure}, it builds the scene representation through three stage: the Coarse-Grained Scene Graph Construction that captures the macro-scale spatial hierarchy of the scene (Sec.~\ref{sec:Coarse_Grained_Scene_Graph_Representation}); the Fine-Grained Scene Graph Construction that captures the fine-scale realistic representation of the scene (Sec.~\ref{sec:Fine_Grained_Scene_Graph_Representation}); the Traversable Topological Graph Construction that endows robots with high-level planning capabilities in the physically interactive real world (Sec.~\ref{sec:Traversable_Topological_Graph_Representation}).

\subsection{Overview}
\label{sec:Problem_Formulation_Overview}
We extend traditional Hierarchical 3D Scene Graphs~\cite{HOVSG} to support more sophisticated interactive robot navigation tasks. Given RGB-D observations and Poses from a physically interactive scene, we model the environment as a Hierarchical Traversable 3D Scene Graph $\mathcal{G} = (G^{S}, G^{N})$, which explicitly models objects’ interactive properties and maps them onto the lower-level topological graph. Specifically, $G^{S}$ denotes the multi-scale hierarchical structural representation of the scene, consisting of two complementary levels: \ding{192} the Coarse-Grained Scene Graph $G^{S}_{\text{coarse}}$, which captures the macro-scale spatial organization across building, floor, and room hierarchies, represented by $V_{\text{coarse}} = \{v^{b}, v^{f}, (v^{r}, \phi_{\text{sem}})\}$; and \ding{193} the Fine-Grained Scene Graph $G^{S}_{\text{fine}}$, which models the micro-scale representation of the scene at the object level, represented by $V_{\text{fine}} = \{(v^{o}, \phi_{\text{sem}}, \phi_{\text{phy}})\}$, where $\phi_{\text{sem}}$ and $\phi_{\text{phy}}$ denote the semantic and interactivity attributes, respectively. \(G^{N}\) denotes the Traversable Navigation Topological Graph, which dynamically models the integration of traversability and interactivity within the navigable regions of the environment, represented by $V_{\text{nav}} = \{(v^{n}, \phi_{\text{free}})\} \cup \{(v^{n}, \phi_{\text{trav}})\}$ where $\phi_{\text{free}}$ and $\phi_{\text{trav}}$ denote the static free-space and the interactive regions associated with movable obstacles, respectively.

\subsection{Coarse-Grained Scene Graph Representation}
\label{sec:Coarse_Grained_Scene_Graph_Representation}

\begin{figure}[t]
    \centering
    \includegraphics[width=\linewidth]{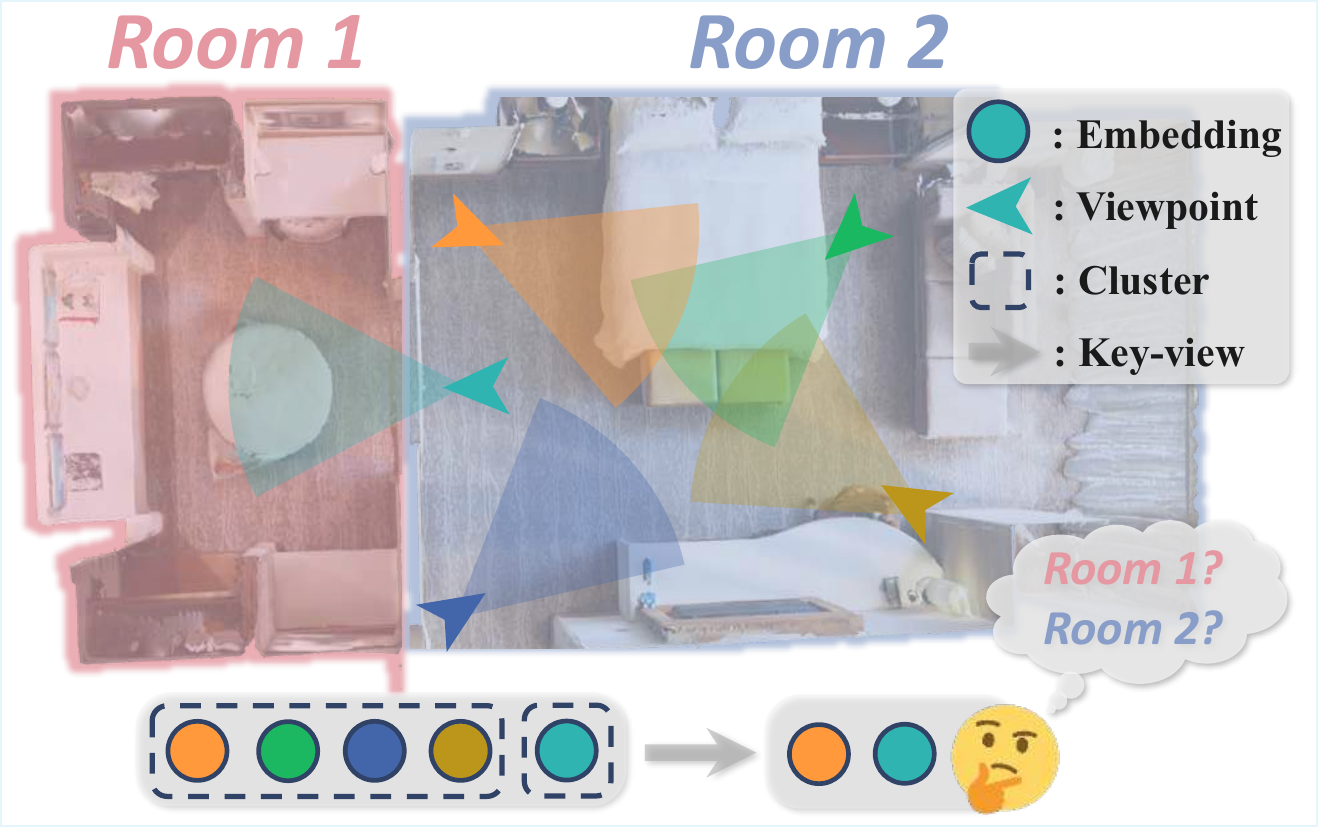}
    \vspace{-0.5cm}
    \caption{
        \textbf{Cross-room Semantic Contamination.} Certain viewpoints capture adjacent regions beyond room boundaries, distorting the intra-room semantic distribution.
    }
    \label{fig:cross_room_contamination}
    \vspace{-0.2cm}
\end{figure}

The Coarse-Grained Scene Graph provides a macro-level abstraction of indoor environments by organizing the scene into hierarchical building-floor-room structures, which establish global structural priors and semantic context essential for high-level reasoning and planning. As shown in Fig.~\ref{fig:main_figure}(a), the structure is constructed through a geometric decomposition of indoor spaces into floor and room components using {{statistics}}-based and {{watershed}}-based approaches\cite{HOVSG, OVIOGW3DHSG, Point2Graph, HiDynaGraph}. This process ensures a well-defined spatial topology that captures the hierarchical organization of large-scale indoor environments (see Appendix \textcolor{cvprblue}{1}).
Subsequently, each room node is endowed with a semantic representation to capture its contextual characteristics within the environment. 

Most existing approaches use K-means-based key-frame selection for room-level feature aggregation, which often introduces cross-room semantic contamination. As shown in Fig.~\ref{fig:cross_room_contamination}, this causes viewpoints near room boundaries to inadvertently capture adjacent spaces, leading to mixed semantics and degraded room embeddings.
To address this issue, we propose a \textbf{Visibility Purification Strategy} that applies visibility-guided weighting to suppress cross-room interference and ensure room-consistent representation.


\textbf{Visibility Purification Strategy} maximizes intra-room coverage diversity while minimizing cross-room interference. 
As illustrated in Fig.~\ref{fig:main_figure}(b), we first perform visibility-based {{weighing}} for each camera view within the room. For camera view $i$ within room $j$, we reconstruct its corresponding 3D observation from the depth map and camera pose. The reconstructed point cloud $\smash{\mathcal{P}^i_{\text{pose}}}$ and the room point cloud $\smash{\mathcal{P}^j_{\text{room}}}$ are then projected onto a unified 2D occupancy grid representing the spatial layout of the room. The proportion of the grid area covered by the projected view indicates how much of the room is visible from that viewpoint, which we define as the visibility weight $w_i$. Subsequently, we perform weighted K-means {{clustering}} on the CLIP embeddings ${\mathbf{f}_i}$ of all images associated with the room, which can be represented as:
\begin{equation}
\min_{\{\boldsymbol{\mu}_k\}_{k=1}^{K}}
\sum_{i=1}^{N} 
w_i\, 
\left\|
\mathbf{f}_i - \boldsymbol{\mu}_{\pi(i)}
\right\|_2^2
\label{eq:weighted_kmeans}
\end{equation}
where $\boldsymbol{\mu}_k$ denotes the centroid of the $k$-th cluster and $\pi(i)$ is the cluster assignment of image $i$. This method prioritizes views with broader spatial coverage while suppressing cross-room interference. Feature-space compactness ensures that semantically coherent views are grouped together, while different observations remain well separated, thus preserving intra-room diversity. The embedding closest to each centroid is selected as its representative, and all representatives are merged by visibility-weighted aggregation to obtain a compact and semantically balanced room representation.


\subsection{Fine-Grained Scene Graph Representation}\label{sec:Fine_Grained_Scene_Graph_Representation}

The Fine-Grained Scene Graph Representation captures detailed geometric structures and localized semantic cues to construct an accurate and realistic indoor scene.
Previous approaches commonly follow a 2D-driven paradigm, projecting dense instance masks from SAM~\cite{sam} into 3D space and merging them by semantic similarity to form object-level nodes. However, this 2D-centric formulation constrains spatial perception to local projections, causing geometric inconsistency, semantic ambiguity, and fragmented object representations.
To construct faithful object representations, we introduce a \textbf{Topological Clustering Strategy} that leverages the global continuity and structural integrity of 3D topology to aggregate geometrically and semantically coherent regions into complete object nodes, while simultaneously enhancing their semantic fidelity by recovering locally missing information and integrating global contextual cues.

\textbf{Topological Clustering Strategy} leverages the global continuity and structural integrity of 3D topology to cluster geometrically connected regions with consistent semantics into complete object-level nodes, effectively mitigating the fragmentation caused by discrete 2D viewpoints.
As shown in Fig.~\ref{fig:main_figure}(c), we first build a {{superpoint graph}} as the structural backbone for topological clustering, where nodes represent locally coherent regions and edges encode geometric adjacency and contextual relationships within each room.
Topological clustering begins by partitioning the input point cloud $\mathcal{P}_{\text{room}}$ into superpoints $\mathcal{S}=\{S_k\}_{k=1}^{M}$ following GrowSP~\cite{GrowSP}, which jointly considers spatial, normal, and normalized RGB distances among 3D points (see Appendix \textcolor{cvprblue}{2}).
A locally connected superpoint graph $\mathcal{G}_{\text{sp}}$ is then constructed based on these superpoints:
\begin{equation}
\mathcal{G}_{\text{sp}} =
\big\{\, (S_i, S_j) \,\big|\,
S_i, S_j \in \mathcal{S},\,
1 \le  \mathcal{N}^{(S_i, S_j)} \le r \big\}
\end{equation}
where $\mathcal{N}^{(S_i, S_j)}$ denotes the neighborhood order between the two superpoints $S_i$ and $S_j$. We then perform similarity mapping to estimate the affinity between adjacent superpoints. Each edge is assigned a similarity score reflecting the correspondence of its connected nodes for subsequent graph-based aggregation. For each edge $(S_i, S_j)$, all camera views jointly observing both regions are collected and processed by $\operatorname{SAM}$~\cite{Semantic_SAM} to obtain 2D instance masks. The projected superpoints are used to evaluate joint visibility and semantic consistency, which are aggregated to compute the final similarity $\operatorname{C}_{S_i,S_j}$, defined as:
\begin{equation}
\operatorname{C_{S_i,S_j}}
=
\frac{1}{m}
\sum_{k=1}^{m}
w_{i,j}^{k}
\frac{
\left\langle 
\{ x_{S_i,d}^{k} \}_{d=1}^{n}, 
\{ x_{S_j,d}^{k} \}_{d=1}^{n}
\right\rangle
}{
\left\| \{ x_{S_i,d}^{k} \}_{d=1}^{n} \right\|_2 \,
\left\| \{ x_{S_j,d}^{k} \}_{d=1}^{n} \right\|_2
}
\end{equation}
where $n$ denotes the number of instances within the $k$-th view $\mathcal{M}_k$, $\{ x_{S_i,d}^{k} \}_{d=1}^{n}$ represents the feature distribution of superpoint $S_i$ over the 2D instance mask in the $k$-th view, and $w_{i,j}^{k}$ indicates the joint visibility of superpoints $S_i$ and $S_j$ in the $k$-th view, which is defined as:
\begin{equation}
w_{i,j}^{k}
=
\frac{
|S_i^{k}|_{\mathrm{vis}}\,
|S_j^{k}|_{\mathrm{vis}}
}{
|S_i|\,|S_j|}
\end{equation}
where $|S_i^{k}|_{\mathrm{vis}}$ and $|S_i|$ denote the visible and total pixel counts of superpoints $S_i$, respectively. Finally, {topological merging} operates on the constructed superpoint graph to partition object nodes.~Following a progressive growing scheme~\cite{SAI3D,SAM2Object}, similarity-based clustering is executed in multiple stages with gradually relaxed thresholds, allowing small coherent regions to merge first and larger structures to form adaptively.
This dynamic process adjusts merging sensitivity based on connectivity confidence, yielding coherent and robust object-level segmentation.
To further ensure semantic completeness and contextual coherence, as shown in Fig.~\ref{fig:main_figure}(d), we refine the merged object nodes by recovering locally missing semantic cues from image-guided observations and reinforcing global context through multi-scale feature aggregation. 
These complementary cues are then used to assign semantically complete and contextually coherent descriptions to the merged object nodes, yielding robust object-level representations even under imperfect observations.

\subsection{Traversable Topological Graph Representation}

\begin{figure}[t]
    \centering
    \includegraphics[width=\linewidth]{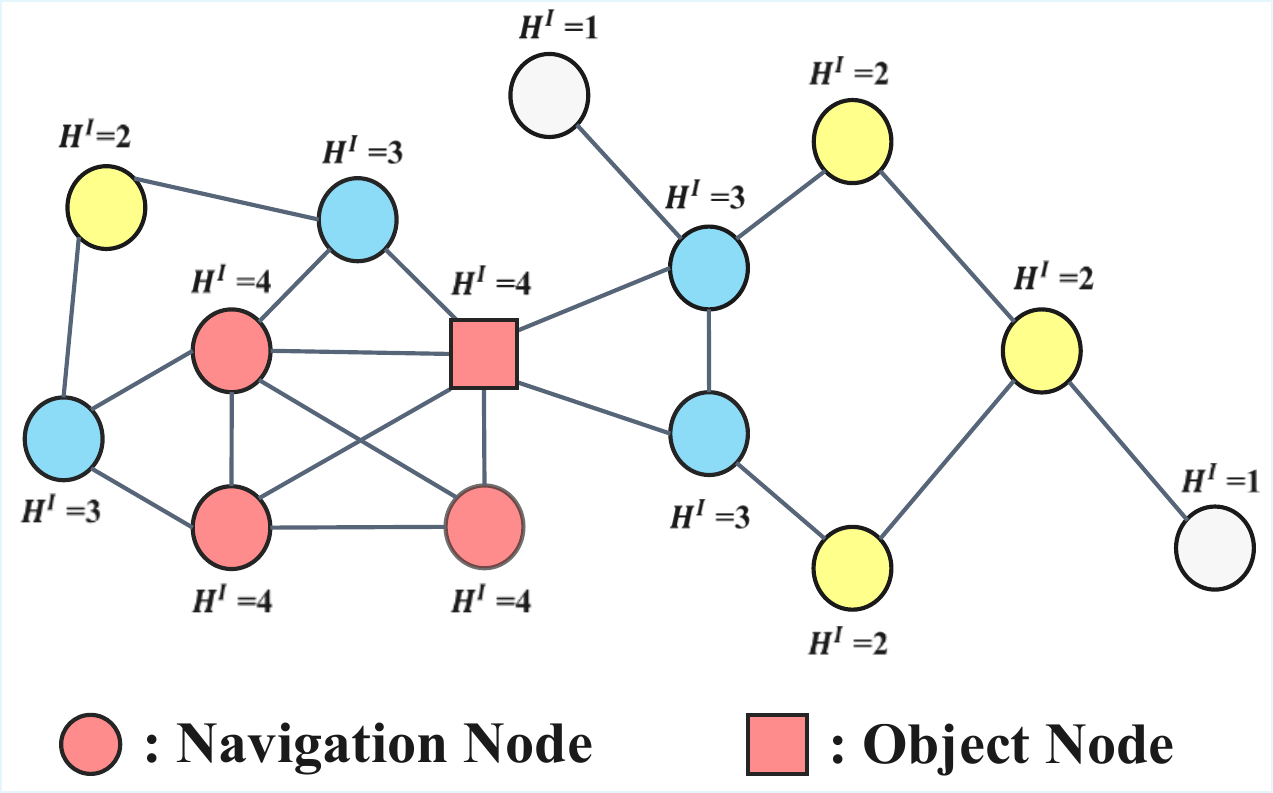}
    \vspace{-0.5cm}
    \caption{
        \textbf{Hierarchical Graph Decomposition.} Object nodes are inserted into the navigation graph, and evaluates their influence to derive hierarchical layers, enabling principled assessment of each candidate object's contribution to navigation efficiency.
    }
    \label{fig:KSIF}
    \vspace{-0.2cm}
\end{figure}


\label{sec:Traversable_Topological_Graph_Representation}
Humans navigate complex environments by interacting with movable objects to create traversable pathways, where navigable space is inherently dynamic and defined by object movability rather than static geometry.
Conventional topological representations~\cite{VoroNav} model free space only relative to static obstacles, thus failing to capture such interactive dynamics.
To overcome this limitation, we introduce the \textbf{Traversability Update Strategy}, which builds upon the Voronoi-based navigation graph (see Appendix \textcolor{cvprblue}{3}) and dynamically integrates object movability into the topological structure, enabling adaptive and human-like navigation in real-world scenes.


\textbf{Traversability Update Strategy} identifies objects with interactive movability and dynamically integrates them into the topological navigation graph, enabling traversable paths that adapt to environmental interactions.
Rather than defining movability purely by physical properties, our strategy adopts a functional and efficiency-driven perspective: an object is considered movable only if its manipulation significantly improves navigational efficiency.
Accordingly, as shown in Fig.~\ref{fig:main_figure}(e), movable obstacle recognition is formulated as a global efficiency estimation problem on the topological graph. This filtering excludes objects whose interactions contribute little to navigation improvement.
To quantify this process, we used the K-Shell iteration factor~\cite{CSRM,KSIF} to identify key nodes, thereby performing hierarchical decomposition of the navigation topology when inserting candidate objects (shown in Fig.~\ref{fig:KSIF}) and calculating efficiency metrics:
\begin{equation}
\mathrm{KS^{IF}_i} = \mathrm{k^s_i \left(1 + \frac{H^I_i}{\bar{H}^I}\right)} \, k_i 
+ \sum_{j \in \Gamma_i} \mathrm{k^s_j \left(1 + \frac{H^I_j}{\bar{H}^I}\right)} \, k_j
\label{eq:ksif}
\end{equation}
This formulation produces a topological efficiency score for each node, ranking objects by their $\mathrm{KS^{IF}}$ values to assess their contribution to navigational efficiency (see Appendix \textcolor{cvprblue}{4} for algorithmic details).
Through \textit{Efficiency Filtering}, objects with scores below the threshold~$\delta$ are considered immovable, while those above it are identified as movable in terms of efficiency.
To further ensure interaction feasibility, \textit{Semantic Filtering} leverages a vision–language model to refine these candidates from semantic and physical perspectives (see Appendix \textcolor{cvprblue}{5}).
Finally, a \textit{Topological Update} integrates the confirmed movable obstacles into the Voronoi-based navigation graph by inserting them as interactive nodes and connecting them through distance-adaptive, visibility-checked edges, yielding a compact yet fully traversable topology.

\section{Experiments}
In this section, we conduct extensive experiments to validate our proposed HERO in terms of its capability and feasibility. We first evaluate the structural accuracy and robustness of the constructed graphs (Sec.~\ref{sec:Evaluation_on_Scene_Representation}). Subsequently, we assess our method’s capability in spatial reasoning (Sec.~\ref{sec:Evaluation_on_3D_Visual_Grounding_Task}), 
and navigation among movable obstacles (Sec.~\ref{sec:Evaluation_on_Interactive_Navigation_Task}) in complex 3D environments. Finally, we perform a series of ablation studies to analyze the contribution of each core component (Sec.~\ref{sec:Ablation_Study}).

\subsection{Evaluation on Scene Representation}
\label{sec:Evaluation_on_Scene_Representation}

\begin{table}[t]
    \centering
    \caption{Evaluations on the validation split of ScanNetV2~\cite{ScanNetV2}. We use \textbf{bold} and \underline{underline} to denote the first and second best performance respectively.}
    \vspace{-0.2cm}
    \resizebox{\linewidth}{!}{
    \begin{tabular}{c|c|cc|cccc}
        \toprule 
        \textbf{Method} & \textbf{Venue} & \textbf{Ins.} & \textbf{Sem.} & \textbf{mIoU} & \textbf{F-mIoU} & \textbf{mAcc} & \textbf{mAP}
        \\
        \midrule\midrule
        \rowcolor{gray!15}
        \multicolumn{8}{l}{\textbf{3D Segmentation}}
        \\
        GrowSP~\cite{GrowSP} & CVPR'23 & \ding{55} & \ding{51} & 25.4 & - & 44.2 & - 
        \\
        Part2Object~\cite{Part2Object} & ECCV'24 & \ding{51} & \ding{55} & - & - & - & 12.6 
        \\
        LogoSP~\cite{LogoSP} & CVPR'25 & \ding{55} & \ding{51} & \textbf{35.8} & - & \underline{50.8} & - 
        \\
        \midrule
        \rowcolor{gray!15}
        \multicolumn{8}{l}{\textbf{3D Scene Graphs}}
        \\
        ConceptFusion~\cite{ConceptFusion} & RSS’23 & \ding{51} & \ding{51} & 11.0 & 12.0 & 21.0 & 5.0
        \\
        ConceptGraph~\cite{Conceptgraphs} & CVPR’24 & \ding{51} & \ding{51} & 16.0 & 20.0 & 28.0 & 6.6
        \\
        HOV-SG~\cite{HOVSG} & RSS’24 & \ding{51} & \ding{51} & 22.2 & 30.3 & 43.1 & 9.7 
        \\
        \textbf{HERO} & \textbf{Ours} & \ding{51} & \ding{51} & \underline{28.4} & \textbf{37.5} & \textbf{56.4} & \textbf{14.1}
        \\
        \bottomrule
    \end{tabular}}
    \vspace{-0.2cm}
    \label{tab:tab1}
\end{table}

We assess structural accuracy and semantic consistency from two complementary perspectives: instance segmentation, which measures the completeness of the nodes independent of semantic categories, and semantic segmentation, which evaluates the fidelity of the semantically annotated nodes. We conduct these evaluations on 100 scenes from the validation split of the richly annotated ScanNetV2~\cite{ScanNetV2} dataset, which comprises hundreds of 3D reconstructed indoor scenes across diverse environments such as offices, hotels, and libraries. We employ standard metrics, including mean Intersection-over-Union (mIoU), Frequency-weighted mean Intersection-over-Union (F-mIoU), and mean class Accuracy (mAcc) for semantic segmentation, and mean Average Precision (mAP) metric for instance segmentation. Detailed experimental settings are provided in Appendix \textcolor{cvprblue}{6}.

As shown in Table~\ref{tab:tab1}, our results highlight the advantages of HERO in producing more faithful and realistic scene representations. Notably, HERO surpasses all 3DSG methods. Compared with the strong baseline, HOV-SG~\cite{HOVSG}, it achieves dramatic improvements of 6.2\% in mIoU, 7.2\% in F-mIoU, 13.3\% in mAcc, and 4.4\% in mAP. Beyond 3DSG baselines, HERO also demonstrates strong competitiveness when compared with task-specific zero-shot 3D segmentation methods. Although it is designed as a unified representation rather than a segmentation-only model, HERO achieves higher mAcc than the latest semantic segmentation approach LogoSP~\cite{LogoSP} (+5.6\%) and outperforms the instance segmentation method Part2Object~\cite{Part2Object} in mAP (+1.5\%). This indicates that our approach provides both semantically discriminative and instance-complete object representations, offering a more consistent and expressive scene abstraction even than methods specialized for a single task.

\subsection{Evaluation on 3D Visual Grounding Task}\label{sec:Evaluation_on_3D_Visual_Grounding_Task}
\begin{table}[t]
    \centering
    \caption{ Evaluations of 3D Visual Grounding on ScanRefer~\cite{ScanRefer} validation set.}
    \vspace{-0.2cm}
    \resizebox{\linewidth}{!}{
    \begin{tabular}{c|c|c|cc}
        \toprule
        \textbf{Method} & \textbf{Venue} & \textbf{Agent}  & \textbf{Acc@$\mathbf{0.25}$} & \textbf{Acc@$\mathbf{0.5}$} 
        \\
        \midrule\midrule
        \rowcolor{gray!15}
        \multicolumn{5}{l}{\textbf{3D Visual Grounding}}
        \\
        OpenScene~\cite{OpenScene} & CVPR’23 & CLIP & 13.2 & 6.5 \\
        ZSVG3D~\cite{ZSVG3D} & CVPR’24 & GPT-4 turbo & 36.4 & 32.7 \\
        SeeGround~\cite{SeeGround} & CVPR’25 & Qwen2-VL-72b & \underline{44.1} & \underline{39.4} \\
        \midrule
        \rowcolor{gray!15}
        \multicolumn{5}{l}{\textbf{3D Scene Graphs}} 
        \\
        ConceptGraphs \cite{Conceptgraphs} & ICRA’24 & CLIP & 14.9 & 6.4 \\
        HOV-SG \cite{HOVSG} & RSS’24 & CLIP & 16.4 & 7.3 \\

        \textbf{HERO}  & \textbf{Ours} & \textbf{CLIP} & \textbf{58.3} & \textbf{43.7} \\

        \bottomrule
    \end{tabular}
    }
    \vspace{-0.2cm}
    \label{tab:tab3}
\end{table}

\begin{table*}[t]
    \centering
    \caption{Evaluation of Interactive Navigation Tasks. We conduct a comprehensive evaluation on 160 tasks across 8 complex indoor environments, where most tasks require interacting with movable obstacles to establish feasible navigation routes. We highlight the key metrics using color annotations.
    }
    \vspace{-0.2cm}
    \resizebox{\linewidth}{!}{
    \begin{tabular}{c|ccc|cccc|cccc}
        \toprule
        \multirow{2}{*}{\textbf{~~ID~~}} & 
        \multirow{2}{*}{\textbf{Blocking}} &
        \multirow{2}{*}{\textbf{\# Movable}} & 
        \multirow{2}{*}{\textbf{\# Tasks}} &
        \multicolumn{4}{c|}{\textbf{Baseline (w/o Interaction)}} & 
        \multicolumn{4}{c}{\textbf{HERO (w/ Interaction)}} 
        \\
        & & & & \textbf{~~PL$\downarrow$~~} & \textbf{~~NE$\downarrow$~~} & \textbf{~~SPL$\uparrow$~~} & \textbf{~~SR$\uparrow$~~} & \textbf{~~PL$\downarrow$~~} & \textbf{~~NE$\downarrow$~~} & \textbf{~~SPL$\uparrow$~~} & \textbf{~~SR$\uparrow$~~} 
        \\
        \midrule\midrule
        
        1 & \ding{55} & 4 & 20 & \cellcolor{yellow!10} 19.0 &   1.2 & \cellcolor{yellow!10}40.6 & 80.0 & \cellcolor{yellow!10} 13.1 & 1.4 & \cellcolor{yellow!10}72.1 & 100.0\\
        2 & \ding{55} & 2 & 20 & \cellcolor{yellow!10} 20.6 &   5.3 & \cellcolor{yellow!10}28.8 & 40.0  & \cellcolor{yellow!10} 14.6 & 0.4 & \cellcolor{yellow!10}73.9 & 100.0\\
        3 & \ding{55} & 5 & 25 & \cellcolor{yellow!10} 39.0 &   3.1 & \cellcolor{yellow!10}45.1 & 80.0 & \cellcolor{yellow!10} 23.4 & 1.2 & \cellcolor{yellow!10}75.2 & 100.0\\

        \midrule
        4 & \ding{51} & 3 & 20 &  9.2 & \cellcolor{orange!10} 9.2 &  3.3 & \cellcolor{orange!10} 5.0 &   9.2 & \cellcolor{orange!10} 3.5 &  64.5 & \cellcolor{orange!10} 85.0\\
        5 & \ding{51} & 3 & 20 &  5.4 &  \cellcolor{orange!10} 7.9 &  5.2 & \cellcolor{orange!10} 10.0 &   5.4 & \cellcolor{orange!10} 0.8 &  74.0 & \cellcolor{orange!10} 95.0 \\
        6 & \ding{51} & 2 & 15 &  11.0 &  \cellcolor{orange!10} 6.7 &  13.4 & \cellcolor{orange!10} 20.0 &   10.7 & \cellcolor{orange!10} 1.1 &  65.3 & \cellcolor{orange!10} 93.3\\

        \midrule
        7 & \ding{51}/\ding{55} & 3 & 20 &  \cellcolor{gray!10}17.4 &   \cellcolor{gray!10}6.7 &  \cellcolor{gray!10}43.3 & \cellcolor{gray!10}65.0  &  \cellcolor{gray!10}17.6 & \cellcolor{gray!10}0.6 &  \cellcolor{gray!10}68.9 & \cellcolor{gray!10}95.0\\
        8 & \ding{51}/\ding{55} & 4 & 20 &  \cellcolor{gray!10}13.2 &   \cellcolor{gray!10}1.7 &  \cellcolor{gray!10}49.6 & \cellcolor{gray!10}75.0  &  \cellcolor{gray!10}11.2 & \cellcolor{gray!10}1.0 &  \cellcolor{gray!10}71.1 & \cellcolor{gray!10}95.0\\

        \bottomrule
    \end{tabular}}
    \vspace{-0.2cm}
    \label{tab:tab4}
\end{table*}
3D Visual Grounding (3DVG) focuses on localizing assigned objects within 3D scenes using natural language descriptions, providing a direct evaluation of our method’s capability to integrate linguistic comprehension with spatial reasoning in cluttered and diverse 3D environments. We evaluate our approach on the ScanRefer~\cite{ScanRefer} benchmark, which offers a large collection of natural language expressions paired with richly annotated indoor scenes. Following~\cite{ZSVG3D}, our experiments are conducted on 100 validation scenes, encompassing approximately 7000 grounding queries and report Acc@0.25 and Acc@0.5, which denote the percentage of samples where the predicted bounding box has an IoU greater than 0.25 or 0.5 with the ground truth. Detailed experimental settings are provided in Appendix \textcolor{cvprblue}{7}.


As shown in Table~\ref{tab:tab3}, HERO demonstrates strong language–scene alignment capability and robust cross-modal retrieval performance, despite not being tailored specifically for grounding. This reflects the semantic completeness and spatial discriminability of its object-level representations. Compared with 3DSG baselines, our method shows a dramatic improvement. Relative to HOV-SG~\cite{HOVSG}, it boosts Acc@0.25 from 16.4\% to 58.3\% (+41.9\%) and Acc@0.5 from 7.3\% to 43.7\% (+36.4\%), highlighting its superior ability to capture fine-grained semantics required for accurate localization. 
Moreover, HERO achieves competitiveness even against dedicated zero-shot 3DVG models. It surpasses OpenScene~\cite{OpenScene} by 45.1\% (Acc@0.25) and 37.2\% (Acc@0.5), and exceeds the performance of LLM-enhanced systems such as SeeGround~\cite{SeeGround}, despite relying only on CLIP~\cite{radford2021learning}. 
These results show HERO's strong generalization, enabling reliable language-guided reasoning across embodied tasks without task-specific designs.



\begin{figure}[t]
    \centering
    \includegraphics[width=\linewidth]{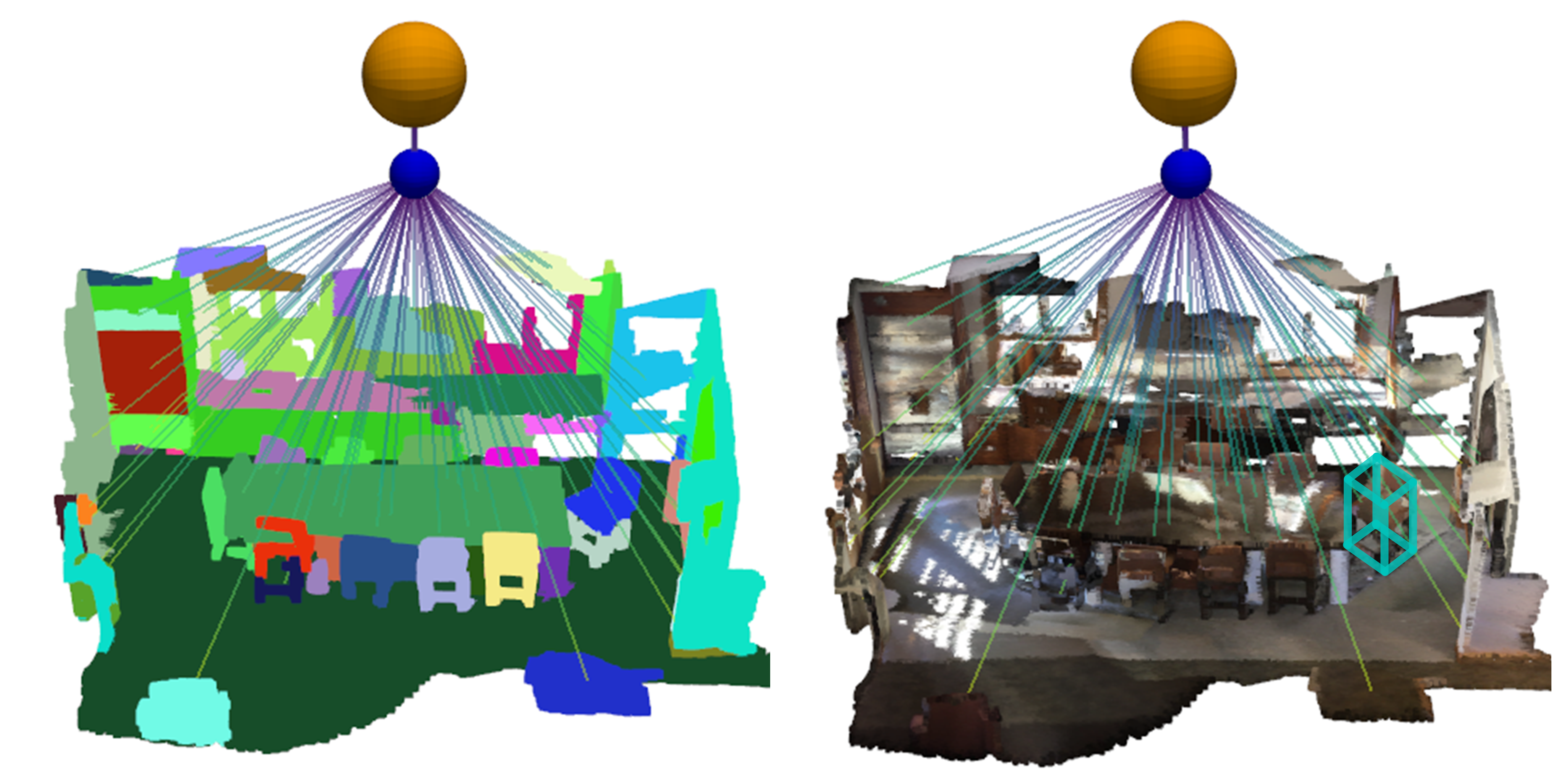}
    \vspace{-0.5cm}
    \caption{
    Visualization of HERO’s structural–semantic segmentation on a ScanNetV2~\cite{ScanNetV2} scene and object localization in a 3D grounding task.
    }
    \label{fig:FIG5}
    \vspace{-0.2cm}
\end{figure}

\subsection{Evaluation on Interactive Navigation Task}\label{sec:Evaluation_on_Interactive_Navigation_Task}
The interactive navigation task highlights the capability of our method to enable efficient and adaptively reachable navigation among movable obstacles in complex, physically realistic environments.
Since existing benchmarks rarely include scenarios that involve movable obstacles in large-scale and structurally complex indoor environments, we construct an augmented version of the HM3D~\cite{HM3D} dataset specifically for evaluation.
Specifically, we select 8 indoor scenes of varying structural complexity and diversity, into which 2 to 5 common movable obstacles are inserted within key traversable regions, partially obstructing critical pathways. Such configurations compel the agent to take considerably excessive detours or even render certain targets unreachable without interaction, establishing physically constrained yet interaction-rich navigation scenarios. In total, we define 160 navigation tasks across these scenes, the majority of which require interactive planning to reach the target.
To highlight the advantages of our method, we compare it with a non-interactive scene graph paradigm by adopting a modified version of HOV-SG~\cite{HOVSG} as the baseline. For quantitative evaluation, we employ several metrics to assess navigation performance, including Path Length (PL), Navigation Error (NE), Success weighted by Path Length (SPL), and Success Rate (SR). Detailed experimental settings are provided in Appendix \textcolor{cvprblue}{8}.

In Table~\ref{tab:tab4}, we highlight the key metrics under different scenario settings using color annotations, which demonstrate that HERO consistently achieves higher efficiency and substantially improved reachability. In scenarios where movable objects do not directly block the traversable space (ID 1–3), as highlighted in yellow, HERO consistently achieves more efficient navigation than the non-interactive baseline. It reduces the average PL from roughly 26.2 m to 17.0 m, an improvement of about 35\%. 
Meanwhile, SPL also improves across all cases, with scene 2 exhibiting the most significant gain, increasing from 28.8 to 73.9 (approximately 2.5$\times$). These results indicate that our method can leverage subtle movability interaction to avoid unnecessary detours. 

Moreover, HERO achieves substantially higher reachability. In scenarios where movable obstacles directly block and divide the traversable space (ID 4–6), as highlighted in orange, HERO exhibits a dramatic improvement in reachability compared with the non-interactive baseline. Both SR and NE are consistently and significantly better across all scenes. Notably, in scene 4, HERO boosts the SR from only 5\% under the baseline to 85\%, representing more than a seventeenfold improvement. Likewise, in scene 5, HERO reduces the NE from 7.9 m to 0.8 m, nearly an order of magnitude decrease. These results indicate that our method enables the agent to interact with obstructing objects and reach targets fundamentally unreachable to conventional non-interactive navigation systems.

Finally, to evaluate performance under non-extreme conditions, we consider mixed scenarios where only a part of the movable obstacles creates blockage (ID 7–8). In these partially obstructed cases, as highlighted in grey, HERO still outperforms the baseline across all key metrics. Although the PL remains similar, likely because the baseline succeeds only on simpler non-interactive routes, HERO achieves lower navigation error and notably higher SPL and SR. These results show that HERO remains effective even when obstruction is partial or inconsistent.

\subsection{Ablation Study}\label{sec:Ablation_Study}

\begin{table}[t]
    \centering
    \caption{Ablation of the Topological Clustering Strategy. We examine the effects of the structural (Seg.) and semantic (Enc.) components and different visual–language encoders on overall performance.}
    \vspace{-0.2cm}
    \resizebox{\linewidth}{!}{
    \begin{tabular}{c|cc|cc|cc}
        \toprule 
        \multirow{2}{*}{\textbf{}} & \multicolumn{2}{c|}{\textbf{Modules}} & \multicolumn{2}{c|}{\textbf{Encoders}} & \multicolumn{2}{c}{\textbf{Performance}} \\
        \cmidrule(lr){2-7}
        & \textbf{Seg.} & \textbf{Enc.} & \textbf{SigLIP} & \textbf{CLIP} & \textbf{Acc@$\mathbf{0.25}$} & \textbf{Acc@$\mathbf{0.5}$} \\
        \midrule\midrule
        \multirow{4}{*}{\rotatebox{45}{Variants}} 
        & \ding{55} & \ding{51} & \ding{55} & \ding{51} & 10.1 & 1.7 \\
        & \ding{55} & \ding{51} & \ding{51} & \ding{55} & 8.7 & 2.2 \\
        & \ding{51} & \ding{55} & \ding{51} & \ding{55} & 28.9 & 17.7 \\
        & \ding{51} & \ding{55} & \ding{55} & \ding{51} & 31.9 & 20.5 \\
        \midrule
        \multirow{2}{*}{\rotatebox{45}{Ours}} 
        & \ding{51} & \ding{51} & \ding{51} & \ding{55} & 56.5 & 43.0 \\
        & \ding{51} & \ding{51} & \ding{55} & \ding{51} & \textbf{59.8} & \textbf{47.1} \\
        \bottomrule
    \end{tabular}}
    \vspace{-0.3cm}
    \label{tab:tab5}
\end{table}

As the key bridge connecting high-level task objectives with low-level navigation execution, the Fine-Grained Scene Graph (Sec.~\ref{sec:Fine_Grained_Scene_Graph_Representation}) plays a decisive role in our system. To clarify its contribution, we conduct an ablation study on 10 scenes from the ScanRefer~\cite{ScanRefer} benchmark, focusing on the object segmentation and encoding components within the Topological Clustering Strategy. We construct four variant configurations by selectively disabling these modules and substituting them with simplified alternatives, and additionally evaluate their behavior when combined with different semantic encoders~\cite{radford2021learning,siglip}. This setup enables a systematic examination of how changes in fine-grained structural and semantic cues influence the overall performance of the framework. Further implementation details and additional ablation studies are provided in Appendix \textcolor{cvprblue}{9}.

As shown in Table~\ref{tab:tab5}, the Topological Clustering Strategy is crucial for constructing reliable object-level representations. Removing the structural module (Seg.) leads to a severe breakdown in performance, with Acc@0.25 falling from 59.8 to 10.1 and Acc@0.5 from 47.1 to 1.7. Disabling the semantic enrichment module (Enc.) produces a less drastic yet still substantial drop, reducing performance by roughly half. Moreover, although SigLIP~\cite{siglip} offers stronger standalone semantic encoding, replacing CLIP~\cite{radford2021learning} by this in the full configuration consistently decreases performance (from 59.8 and 47.1 to 56.5 and 43.0), indicating that encoder strength alone does not ensure compatibility. Instead, the structural and semantic cues produced by our pipeline align more effectively with CLIP’s feature space. Overall, these results show that both structural and semantic components, together with their compatibility with the chosen encoder, are critical for achieving reliable retrieval.



\section{Conclusion}
This paper presents HERO, a framework for Hierarchical Traversable 3D Scene Graphs that goes beyond static-world assumptions by explicitly modeling structural hierarchy, semantics, and interactive dynamics for navigation among movable obstacles. HERO targets a key weakness of existing navigation pipelines: the scene graphs they used are characterized by noisy and incomplete semantics for downstream decision-making. To address this, our Visibility Purification Strategy suppresses cross-room semantic contamination and yields viewpoint-consistent room representations, while the Topological Clustering Strategy performs geometry-aware and multi-view aggregation to produce object nodes that are both topologically coherent and semantically complete. Built on these refined semantics, our Traversability Update Strategy integrates movable obstacles into the navigation graph via an efficiency-driven formulation of functional movability, redefining traversable regions and enabling more human-like navigation. Extensive experiments on structural segmentation, 3D visual grounding, and interactive navigation show that HERO yields higher-quality semantics and consistently better navigation performance than scene-graph and task-specific baselines.

{
    \small
    \bibliographystyle{ieeenat_fullname}
    \bibliography{main}

@inproceedings{SFM,
  title={Structure-from-motion revisited},
  author={Schonberger, Johannes L and Frahm, Jan-Michael},
  booktitle={Proceedings of the IEEE conference on computer vision and pattern recognition},
  pages={4104--4113},
  year={2016}
}

@inproceedings{SAM2Object,
  title={SAM2Object: Consolidating View Consistency via SAM2 for Zero-Shot 3D Instance Segmentation},
  author={Zhao, Jihuai and Zhuo, Junbao and Chen, Jiansheng and Ma, Huimin},
  booktitle={Proceedings of the IEEE/CVF Conference on Computer Vision and Pattern Recognition},
  pages={19325-19334},
  year={2025},
}

@inproceedings{VCCS,
  title={Voxel Cloud Connectivity Segmentation - Supervoxels for Point Clouds},
  author={Papon, Jeremie and Abramov, Alexey and Schoeler, Markus and Wörgötter, Florentin},
  booktitle={Proceedings of the IEEE/CVF Conference on Computer Vision and Pattern Recognition},
  pages={2027-2034},
  year={2013},
}

@inproceedings{ScanRefer,
  title={ScanRefer: 3D Object Localization in RGB-D Scans using Natural Language},
  author={Chen, Dave Zhenyu and Chang, Angel X and Nie{\ss}ner, Matthias},
  booktitle={European Conference on Computer Vision},
  pages={202-221},
  year={2020},
}

@inproceedings{Conceptgraphs,
  title={Conceptgraphs: Open-vocabulary 3d scene graphs for perception and planning},
  author={Gu, Qiao and Kuwajerwala, Ali and Morin, Sacha and Jatavallabhula, Krishna Murthy and Sen, Bipasha and Agarwal, Aditya and Rivera, Corban and Paul, William and Ellis, Kirsty and Chellappa, Rama and others},
  booktitle={IEEE International Conference on Robotics and Automation},
  pages={5021--5028},
  year={2024},
}

@inproceedings{INIRWTOULLAVLM,
  title={Interactive Navigation in Environments with Traversable Obstacles Using Large Language and Vision-Language Models},
  author={Zhen Zhang and Anran Lin and Chun Wai Wong and Xiangyu Chu and Qi Dou and K. W. Samuel Au},
  booktitle={IEEE International Conference on Robotics and Automation},
  pages={7867-7873},
  year={2024},
}

@inproceedings{ZSVG3D,
  title={Visual Programming for Zero-shot Open-Vocabulary 3D Visual Grounding},
  author={Zhihao Yuan and Jinke Ren and Chun-Mei Feng and Hengshuang Zhao and Shuguang Cui and Zhen Li},
  booktitle={Proceedings of the IEEE/CVF Conference on Computer Vision and Pattern Recognition},
  pages={20623-20633},
  year={2024}
}

@inproceedings{GrowSP,
  title={GrowSP: Unsupervised Semantic Segmentation of 3D Point Clouds},
  author={Zihui Zhang and Bo Yang and Bing Wang and Bo Li},
  booktitle={Proceedings of the IEEE/CVF Conference on Computer Vision and Pattern Recognition},
  pages={17619-17629},
  year={2023}
}

@inproceedings{ScanNetV2,
    title={ScanNet: Richly-annotated 3D Reconstructions of Indoor Scenes},
    author={Dai, Angela and Chang, Angel X. and Savva, Manolis and Halber, Maciej and Funkhouser, Thomas and Nie{\ss}ner, Matthias},
    booktitle = {Proceedings of the IEEE/CVF Conference on Computer Vision and Pattern Recognition},
    year = {2017}
}

@inproceedings{Semantic_SAM,
  title={Semantic-SAM: Segment and Recognize Anything at Any Granularity},
  author={Li, Feng and Zhang, Hao and Sun, Peize and Zou, Xueyan and Liu, Shilong and Yang, Jianwei and Li, Chunyuan and Zhang, Lei and Gao, Jianfeng},
  booktitle={European Conference on Computer Vision},
  pages={467-484},
  year={2023}
}

@inproceedings{Part2Object,
  title={Part2Object: Hierarchical Unsupervised 3D Instance Segmentation},
  author={Shi, Cheng and Zhang, Yulin and Yang, Bin and Tang, Jiajin and Ma, Yuexin and Yang, Sibei},
  booktitle={European Conference on Computer Vision},
  pages={1-18},
  year={2024}
}

@inproceedings{OpenScene,
  title={OpenScene: 3D Scene Understanding with Open Vocabularies},
  author={Peng, Songyou and Genova, Kyle and Jiang, Chiyu "Max" and Tagliasacchi, Andrea and Pollefeys, Marc and Funkhouser, Thomas},
  booktitle={Proceedings of the IEEE/CVF Conference on Computer Vision and Pattern Recognition},
  pages={815-824},
  year={2023}
}

@inproceedings{NAMOWOLUPS,
  title={Navigation Among Movable Obstacles with Object Localization using Photorealistic Simulation},
  author={Ellis, Kirsty and Zhang, Henry and Stoyanov, Danail and Kanoulas, Dimitrios},
  booktitle={IEEE/RSJ International Conference on Intelligent Robots and Systems},
  pages={1711--1716},
  year={2022}
}

@inproceedings{IMAGE2MASS,
  title={image2mass: Estimating the Mass of an Object from Its Image},
  author={Standley, Trevor and Sener, Ozan and Chen, Dawn and Savarese, Silvio},
  booktitle={Proceedings of the Conference on Robot Learning},
  pages={324--333},
  year={2017}
}

@inproceedings{ABMRNAMO,
  title={Affordance-Based Mobile Robot Navigation Among Movable Obstacles},
  author={Maozhen Wang and Rui Luo and Aykut Ozgun Onol and Taskin Padir},
  booktitle={IEEE/RSJ International Conference on Intelligent Robots and Systems},
  pages={2734-2740},
  year={2020}
}

@inproceedings{IFAR,
  title={Interactive-FAR:Interactive, Fast and Adaptable Routing for Navigation Among Movable Obstacles in Complex Unknown Environments},
  author={Botao He and Guofei Chen and Wenshan Wang and Ji Zhang and Cornelia Fermuller and Yiannis Aloimonos},
  booktitle={IEEE/RSJ International Conference on Intelligent Robots and Systems},
  pages={5402-5409},
  year={2024}
}

@inproceedings{3DSGs,
  title={3D Scene Graph: A Structure for Unified Semantics, 3D Space, and Camera},
  author={Armeni, Iro and He, Zhi-Yang and Gwak, JunYoung and Zamir, Amir R and Fischer, Martin and Malik, Jitendra and Savarese, Silvio},
  booktitle={Proceedings of the IEEE International Conference on Computer Vision},
  pages={5664--5673},
  year={2019}
}

@inproceedings{HOVSG,
  title={Hierarchical Open-Vocabulary 3D Scene Graphs for Language-Grounded Robot Navigation},
  author={Abdelrhman Werby and Chenguang Huang and Martin Büchner and Abhinav Valada and Wolfram Burgard},
  booktitle={Proceedings of Robotics: Science and Systems},
  year={2024}
}

@inproceedings{ConceptFusion,
  title={ConceptFusion: Open-set Multimodal 3D Mapping},
  author={Jatavallabhula, {Krishna Murthy} and Kuwajerwala, Alihusein and Gu, Qiao and Omama, Mohd and Chen, Tao and Li, Shuang and Iyer, Ganesh and Saryazdi, Soroush and Keetha, Nikhil and Tewari, Ayush and Tenenbaum, {Joshua B.} and {de Melo}, {Celso Miguel} and Krishna, Madhava and Paull, Liam and Shkurti, Florian and Torralba, Antonio},
  booktitle={Proceedings of Robotics: Science and Systems},
  year={2023}
}

@inproceedings{OpenGraph,
  title={OpenGraph: Towards Open Graph Foundation Models},
  author={Xia, Lianghao and Kao, Ben and Huang, Chao},
  booktitle={Empirical Methods in Natural Language Processing},
  year={2024}
}

@inproceedings{SeeGround,
  title     = {SeeGround: See and Ground for Zero-Shot Open-Vocabulary 3D Visual Grounding},
  author    = {Rong Li and Shijie Li and Lingdong Kong and Xulei Yang and Junwei Liang},
  booktitle = {Proceedings of the IEEE/CVF Conference on Computer Vision and Pattern Recognition},
  year      = {2025},
}

@inproceedings{SceneGraphFusion,
        title={SceneGraphFusion: Incremental 3D Scene Graph Prediction from RGB-D Sequences},
        author={Shun-Cheng Wu and Johanna Wald and Keisuke Tateno and Nassir Navab and Federico Tombari},
        booktitle={Proceedings of the IEEE/CVF Conference on Computer Vision and Pattern Recognition},
        pages={7515-7525},
        year={2021},
      }

@inproceedings{ASG3DSG,
    title={A Survey on 3D Scene Graphs: Definition, Generation and Application},
    author={Bae, Jaewon and Shin, Dongmin and Ko, Kangbeen and Lee, Juchan and Kim, Ue-Hwan},
    booktitle={Robot Intelligence Technology and Applications 7},
    pages={136--147},
    year={2023},
  }

@inproceedings{LogoSP,
        title={LogoSP: Local-global Grouping of Superpoints for Unsupervised Semantic Segmentation of 3D Point Clouds},
        author={Zihui Zhang and Weisheng Dai and Hongtao Wen and Bo Yang},
        booktitle={Proceedings of the IEEE/CVF Conference on Computer Vision and Pattern Recognition},
        pages={1374-1384},
        year={2025},
      }

@inproceedings{SAI3D,
        title={SAI3D: Segment Any Instance in 3D Scenes},
        author={Yingda Yin and Yuzheng Liu and Yang Xiao and Daniel Cohen-Or and Jingwei Huang and Baoquan Chen},
        booktitle={Proceedings of the IEEE/CVF Conference on Computer Vision and Pattern Recognition},
        pages={3292-3302},
        year={2024},
      }

@article{NAMO,
  title={NAVIGATION AMONG MOVABLE OBSTACLES: REAL-TIME REASONING IN COMPLEX ENVIRONMENTS},
  author={Stilman, Mike, and James J. Kuffner},
  journal={International Journal of Humanoid Robotics},
  volume={2},
  number = {04},
  pages={479-503},
  year={2005}
}

@article{SWIPEBOT,
  title   = {SwipeBot: DNN-based Autonomous Robot Navigation among Movable Obstacles in Cluttered Environments},
  author  = {Nikolay Zherdev and Mikhail Kurenkov and Kristina Belikova and Dzmitry Tsetserukou},
  journal = {arXiv preprint arXiv:2305.04851},
  year    = {2023}
}

@article{ENAMOUAMMVHPL,
  title   = {Efficient Navigation Among Movable Obstacles using a Mobile Manipulator via Hierarchical Policy Learning},
  author  = {Taegeun Yang and Jiwoo Hwang and Jeil Jeong and Minsung Yoon and Sung-Eui Yoon},
  journal = {arXiv preprint arXiv:2506.15380},
  year    = {2025}
}

@article{JPAPPAMO,
  title   = {Joint Path and Push Planning Among Movable Obstacles},
  author  = {Victor Emeli and Akansel Cosgun},
  journal = {arXiv preprint arXiv:2010.14733},
  year    = {2020}
}

@article{HM3D,
  title   = {Habitat-Matterport 3D Dataset (HM3D): 1000 Large-scale 3D Environments for Embodied AI},
  author  = {Santhosh K. Ramakrishnan and Aaron Gokaslan and Erik Wijmans and Oleksandr Maksymets and Alex Clegg and John Turner and Eric Undersander and Wojciech Galuba and Andrew Westbury and Angel X. Chang and Manolis Savva and Yili Zhao and Dhruv Batra},
  journal = {arXiv preprint arXiv:2109.08238},
  year    = {2021}
}

@article{VoroNav,
  title={VoroNav: Voronoi-based Zero-shot Object Navigation with Large Language Model},
  author={Wu, Pengying and Mu, Yao and Wu, Bingxian and Hou, Yi and Ma, Ji and Zhang, Shanghang and Liu, Chang},
  journal={arXiv preprint arXiv:2401.02695},
  year={2024}
}

@article{NAMOUNC,
  title   = {NAMOUnc: Navigation Among Movable Obstacles with Decision Making on Uncertainty Interval},
  author  = {Kai Zhang and Eric Lucet and Julien Alexandre Dit Sandretto and Shoubin Chen and David Filliat},
  journal = {arXiv preprint arXiv:2509.12723},
  year    = {2025}
}

@article{PTCWMAOBO,
  title   = {Pushing Through Clutter With Movability Awareness of Blocking Obstacles},
  author  = {Joris J. Weeda and Saray Bakker and Gang Chen and Javier Alonso-Mora},
  journal = {arXiv preprint arXiv:2502.20106},
  year    = {2025}
}

@article{AINOQRULLM,
  title   = {Adaptive Interactive Navigation of Quadruped Robots using Large Language Models},
  author  = {Kangjie Zhou and Yao Mu and Haoyang Song and Yi Zeng and Pengying Wu and Han Gao and Chang Liu},
  journal = {arXiv preprint arXiv:2503.22942},
  year    = {2025}
}

@article{FSRVLN,
  title   = {FSR-VLN: Fast and Slow Reasoning for Vision-Language Navigation with Hierarchical Multi-modal Scene Graph},
  author  = {Xiaolin Zhou and Tingyang Xiao and Liu Liu and Yucheng Wang and Maiyue Chen and Xinrui Meng and Xinjie Wang and Wei Feng and Wei Sui and Zhizhong Su},
  journal = {arXiv preprint arXiv:2509.13733},
  year    = {2025}
}

@article{TBNOUODNIUEWMO,
  title={Tactile‐Based Negotiation of Unknown Objects during Navigation in Unstructured Environments with Movable Obstacles},
  author={Armleder, Simon and Dean, Emmanuel and Bergner, Florian and Guadarrama Olvera, Julio Rogelio and Cheng, Gordon},
  journal={Advanced Intelligent Systems},
  volume={6},
  number = {3},
  pages={2300621},
  year={2024}
}

@article{IAO3DSGVLESO,
  title={Indoor and Outdoor 3D Scene Graph Generation via Language-Enabled Spatial Ontologies},
  author={Jared Strader and Nathan Hughes and William Chen and Alberto Speranzon and Luca Carlone},
  journal={IEEE Robotics and Automation Letters},
  volume={9},
  number = {6},
  pages={4886-4893},
  year={2024}
}

@article{Point2Graph,
  title   = {Point2Graph: An End-to-end Point Cloud-based 3D Open-Vocabulary Scene Graph for Robot Navigation},
  author  = {Yifan Xu and Ziming Luo and Qianwei Wang and Vineet Kamat and Carol Menassa},
  journal = {arXiv preprint arXiv:2409.10350},
  year    = {2024}
}

@article{ATSMFGLH,
  title={A Threshold Selection Method from Gray-Level Histograms},
  author={Otsu, Nobuyuki},
  journal={IEEE Transactions on Systems, Man, and Cybernetics},
  volume={9},
  number = {1},
  pages={62-66},
  year={1979}
}

@article{AOOWAIIOSL,
  title={An Overview of Watershed Algorithm Implementations in Open Source Libraries},
  author={Kornilov, Anton S. and Safonov, Ilia V.},
  journal={Journal of Imaging},
  volume={4},
  number = {10},
  pages={123},
  year={2018}
}

@article{OVIOGW3DHSG,
  title   = {Open-Vocabulary Indoor Object Grounding with 3D Hierarchical Scene Graph},
  author  = {Sergey Linok and Gleb Naumov},
  journal = {arXiv preprint arXiv:2507.12123},
  year    = {2025}
}

@article{sam,
  title   = {Segment Anything},
  author  = {Alexander Kirillov and Eric Mintun and Nikhila Ravi and Hanzi Mao and Chloe Rolland and Laura Gustafson and Tete Xiao and Spencer Whitehead and Alexander C. Berg and Wan-Yen Lo and Piotr Dollár and Ross Girshick},
  journal = {arXiv preprint arXiv:2304.02643},
  year    = {2023}
}

@article{HiDynaGraph,
  title   = {Hi-Dyna Graph: Hierarchical Dynamic Scene Graph for Robotic Autonomy in Human-Centric Environments},
  author  = {Jiawei Hou and Xiangyang Xue and Taiping Zeng},
  journal = {arXiv preprint arXiv:2506.00083},
  year    = {2025}
}

@article{NAMO_LLM,
  title   = {NAMO-LLM: Efficient Navigation Among Movable Obstacles with Large Language Model Guidance},
  author  = {Yuqing Zhang and Yiannis Kantaros},
  journal = {arXiv preprint arXiv:2505.04141},
  year    = {2025}
}

@article{3DGIR,
  title   = {3D Scene Graphs in Robotics: A Unified Representation Bridging Geometry, Semantics, and Action.},
  author  = {Iacopo Catalano and Carlos Cueto Zumaya and Julio A Placed and Javier Civera and Wallace Moreira Bessa and Jorge Peña-Queralta},
  journal = {TechRxiv},
  year    = {2025}
}

@article{SLAM,
  title={Past, Present, and Future of Simultaneous Localization And Mapping: Towards the Robust-Perception Age},
  author={C. Cadena and L. Carlone and H. Carrillo and Y. Latif and D. Scaramuzza and J. Neira and I. Reid and J.J. Leonard},
  journal={IEEE Transactions on Robotics},
  volume={32},
  number = {6},
  pages={1309–1332},
  year={2016}
}

@article{KSIF,
  title={Fast ranking influential nodes in complex networks using a k-shell iteration factor},
  author={Zhixiao Wang and Ya Zhao and Jingke Xi and Changjiang Du},
  journal={Physica A: Statistical Mechanics and its Applications},
  volume={461},
  pages={171-181},
  year={2016}
}

@article{CSRM,
  title={Identifying influential nodes through an improved k-shell iteration factor model},
  author={Qing Yang and Yunheng Wang and Senbin Yu and Wenjie Wang},
  journal={Expert Systems with Applications},
  volume={238},
  pages={122077},
  year={2024}
}

@article{K_shell,
  title={Identification of influential spreaders in complex networks},
  author={Kitsak, Maksim and Gallos, Lazaros K. and Havlin, Shlomo and Liljeros, Fredrik and Muchnik, Lev and Stanley, H. Eugene and Makse, Hernán A.},
  journal={Nature Physics},
  volume={6},
  number = {11},
  pages={888-893},
  year={2010}
}

@article{HumanNav,
  title={Early electrophysiological markers of navigational affordances in scenes},
  author={Harel, Assaf and Nador, Jeffery D and Bonner, Michael F and Epstein, Russell A},
  journal={Journal of Cognitive Neuroscience},
  volume={34},
  number={3},
  pages={397--410},
  year={2022},
  publisher={MIT Press One Rogers Street, Cambridge, MA 02142-1209, USA journals-info~…}
}

@article{degree,
  title={Centrality in social networks conceptual clarification},
  author={Linton C. Freeman},
  journal={Social Networks},
  volume={1},
  number={3},
  pages={215-239},
  year={1978},
}

@inproceedings{Parking-SG,
  title={Parking-SG: Open-Vocabulary Hierarchical 3D Scene Graph Representation for Open Parking Environments},
  author={Zhang, Yaowen and Ruan, Yi and Pan, Miaoxin and Yang, Yi and Fu, Mengyin},
  booktitle={2025 IEEE International Conference on Robotics and Automation (ICRA)},
  pages={7291--7297},
  year={2025},
  organization={IEEE}
}

@article{tan2025roboos,
  title={Roboos: A hierarchical embodied framework for cross-embodiment and multi-agent collaboration},
  author={Tan, Huajie and Hao, Xiaoshuai and Chi, Cheng and Lin, Minglan and Lyu, Yaoxu and Cao, Mingyu and Liang, Dong and Chen, Zhuo and Lyu, Mengsi and Peng, Cheng and others},
  journal={arXiv preprint arXiv:2505.03673},
  year={2025}
}

@article{gemini,
      title={Gemini 2.5: Pushing the Frontier with Advanced Reasoning, Multimodality, Long Context, and Next Genegeminiration Agentic Capabilities}, 
      author={Gheorghe Comanici and Eric Bieber and Mike Schaekermann and Ice Pasupat and Noveen Sachdeva and Inderjit Dhillon and Marcel Blistein and Ori Ram and Dan Zhang and Evan Rosen and others},
      journal={arXiv preprint arXiv:2507.06261},
      year={2025}
}

@misc{gpt4o,
      title={GPT-4o System Card}, 
      author={OpenAI and Aaron Hurst and Adam Lerer and Adam P. Goucher and Adam Perelman and Aditya Ramesh and Aidan Clark and AJ Ostrow and Akila Welihinda and Alan Hayes and others},
      year={2024},
      journal={arXiv preprint arXiv:2410.21276}
}

@article{chang2024partnr,
  title={Partnr: A benchmark for planning and reasoning in embodied multi-agent tasks},
  author={Chang, Matthew and Chhablani, Gunjan and Clegg, Alexander and Cote, Mikael Dallaire and Desai, Ruta and Hlavac, Michal and Karashchuk, Vladimir and Krantz, Jacob and Mottaghi, Roozbeh and Parashar, Priyam and others},
  journal={arXiv preprint arXiv:2411.00081},
  year={2024}
}

@article{honerkamp2024language,
  title={Language-grounded dynamic scene graphs for interactive object search with mobile manipulation},
  author={Honerkamp, Daniel and B{\"u}chner, Martin and Despinoy, Fabien and Welschehold, Tim and Valada, Abhinav},
  journal={IEEE Robotics and Automation Letters},
  year={2024},
  publisher={IEEE}
}

@article{yan2025dynamic,
  title={Dynamic open-vocabulary 3d scene graphs for long-term language-guided mobile manipulation},
  author={Yan, Zhijie and Li, Shufei and Wang, Zuoxu and Wu, Lixiu and Wang, Han and Zhu, Jun and Chen, Lijiang and Liu, Jihong},
  journal={IEEE Robotics and Automation Letters},
  year={2025},
  publisher={IEEE}
}

@inproceedings{koch2024open3dsg,
  title={Open3dsg: Open-vocabulary 3d scene graphs from point clouds with queryable objects and open-set relationships},
  author={Koch, Sebastian and Vaskevicius, Narunas and Colosi, Mirco and Hermosilla, Pedro and Ropinski, Timo},
  booktitle={Proceedings of the IEEE/CVF Conference on Computer Vision and Pattern Recognition},
  pages={14183--14193},
  year={2024}
}

@inproceedings{chen2024clip,
  title={Clip-driven open-vocabulary 3d scene graph generation via cross-modality contrastive learning},
  author={Chen, Lianggangxu and Wang, Xuejiao and Lu, Jiale and Lin, Shaohui and Wang, Changbo and He, Gaoqi},
  booktitle={Proceedings of the IEEE/CVF Conference on Computer Vision and Pattern Recognition},
  pages={27863--27873},
  year={2024}
}

@inproceedings{radford2021learning,
  title={Learning transferable visual models from natural language supervision},
  author={Radford, Alec and Kim, Jong Wook and Hallacy, Chris and Ramesh, Aditya and Goh, Gabriel and Agarwal, Sandhini and Sastry, Girish and Askell, Amanda and Mishkin, Pamela and Clark, Jack and others},
  booktitle={International conference on machine learning},
  pages={8748--8763},
  year={2021},
  organization={PmLR}
}

@inproceedings{siglip,
  title={Sigmoid loss for language image pre-training},
  author={Zhai, Xiaohua and Mustafa, Basil and Kolesnikov, Alexander and Beyer, Lucas},
  booktitle={Proceedings of the IEEE/CVF international conference on computer vision},
  pages={11975--11986},
  year={2023}
}

@inproceedings{IGBATMFMRN,
  title={Integrating grid-based and topological maps for mobile robot navigation},
  author={Thrun, Sebastian and B\"{u}, Arno},
  booktitle={Proceedings of the national conference on artificial intelligence},
  pages={944-951},
  year={1996}
}
}

\clearpage
\setcounter{page}{1}
\setcounter{section}{0}
\maketitlesupplementary

\section{Floor and Room Decomposition Details}
\label{sec:Floor_and_Room_Node_Partitioning}

\subsection{Floor Node Partitioning}
Some indoor environments typically consist of several vertically stacked floors that may share similar local appearance but differ significantly in their functional layout and connectivity. Explicit floor partitioning establishes the macro-level structural backbone of the scene graph, enabling better alignment with high-level task while improving both retrieval precision and computational efficiency. 

We recover the multi-floor topology of indoor environments by analyzing the vertical distribution of the global point cloud $\mathcal{P}$~\cite{HOVSG, OVIOGW3DHSG, FSRVLN, Point2Graph, HiDynaGraph}. The vertical geometry is modeled as a continuous mapping from height to point density, which is defined as: 
\begin{equation}
\rho(h) = \sum_{p_i \in \mathcal{P}} \mathbb{I} \left( \left| z_i - h \right| < \tfrac{\Delta h}{2} \right)
\label{eq:height_density}
\end{equation}
where $z_i$ is the height coordinate of point $p_i$, $\Delta h$ is the discretization interval along the gravity axis, and $\mathbb{I}(\cdot)$ denotes the indicator function. 
We discretize the entire height range with $\Delta h = 0.01 \text{m}$ and compute a 1D histogram over all points. Peaks in this histogram correspond to prominent horizontal structures such as floors and ceilings.
To extract these structures reliably, we detect local maxima within a neighborhood of $\pm 0.2 \text{m}$ along the height axis and keep only those whose density exceeds 90\% of the global maximum. This filtering step eliminates weak peaks produced by small furniture or minor architectural components. The retained maxima are then grouped in height space using DBSCAN to merge duplicated responses originating from the same physical slab. Within each cluster, we select the two maxima with the highest densities as the representative structural planes and use them to instantiate a floor node $v^{f}$. Finally, each floor node is connected to the building root node $v^{b}$, establishing a coherent building–floor hierarchy that forms the basis for room partitioning and subsequent fine-grained scene graph construction.

\subsection{Room Node Partitioning}

\begin{figure}[t]
    \centering
    \includegraphics[width=\linewidth]{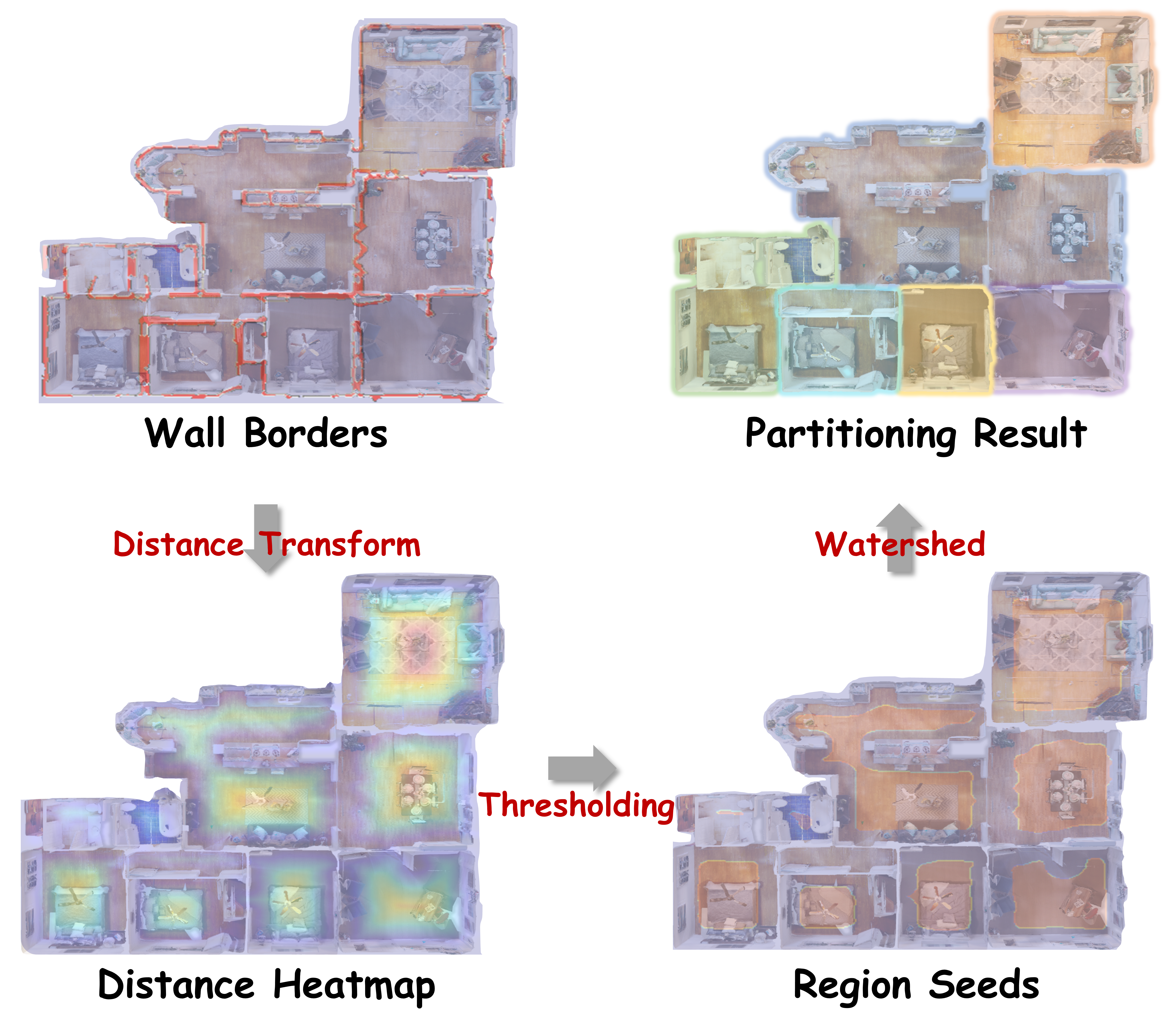}
    \vspace{-0.5cm}
    \caption{\textbf{Room partitioning workflow.} Wall borders are extracted from the BEV occupancy map; a distance transform produces the distance heatmap; region seeds are obtained via adaptive thresholding; and the Watershed algorithm generates the final room partitions.}
    \label{fig:RoomPartitioning}
    \vspace{-0.5cm}
\end{figure}

Indoor spaces on the same floor are typically organized into functionally coherent regions such as bedrooms, kitchens, and offices. Explicit room partitioning therefore provides a mid-level abstraction that bridges the gap between floor-level structure and object-level details, aligns more naturally with high-level task instructions, and improves both retrieval accuracy and computational efficiency by restricting search and reasoning to room-specific subgraphs.

To derive the room-level structure, we first project the floor-specific point cloud $\mathcal{P}_{\text{floor}}$ onto the horizontal plane to obtain a normalized bird’s-eye-view (BEV) occupancy map, where each pixel aggregates the vertical support of all points above it. As illustrated in Fig.~\ref{fig:RoomPartitioning}, we extract a wall border map from the BEV representation by thresholding the occupancy values, which highlights vertically elongated architectural elements such as walls and partitions while suppressing clutter and small objects.
we then perform a distance transform to compute a Euclidean Distance Field (EDF) over the floor plane, where each pixel records its distance to the nearest wall pixel. The resulting distance heatmap captures the free-space geometry shaped by these structural boundaries, with high-valued regions indicating interior areas that naturally serve as candidate room centers. 
These candidate regions are further isolated using Otsu’s adaptive thresholding~\cite{ATSMFGLH}, which determines an optimal threshold $\tau^{*}$ and yields a corresponding set of region seeds.
Using these seeds as initialization markers, we apply the Watershed~\cite{AOOWAIIOSL} algorithm to obtain the final 2D room partitions:
\begin{equation}
\{R_k\} = \operatorname{Watershed}\!\Big(-EDF,\, \mathbb{I}\big(EDF > \tau^{*}\big)\Big)
\label{eq:watershed}
\end{equation}
where each region $R_k$ denotes one room segment on the floor in the BEV domain. 
Each 2D room mask is lifted to 3D by collecting points within its horizontal footprint and floor interval, producing the room-specific point cloud $\mathcal{P}_{\text{room}}$ and the corresponding room node $v^{r}$. Each floor node $v^{f}$ is then connected to its room nodes $(v^{f}, v^{r})$, forming the floor–room hierarchy that underpins subsequent object-level construction and room-aware retrieval.

\section{Superpoint Construction}
\label{sec:Superpoint_Construction}

\begin{figure}[t]
    \centering
    \includegraphics[width=\linewidth]{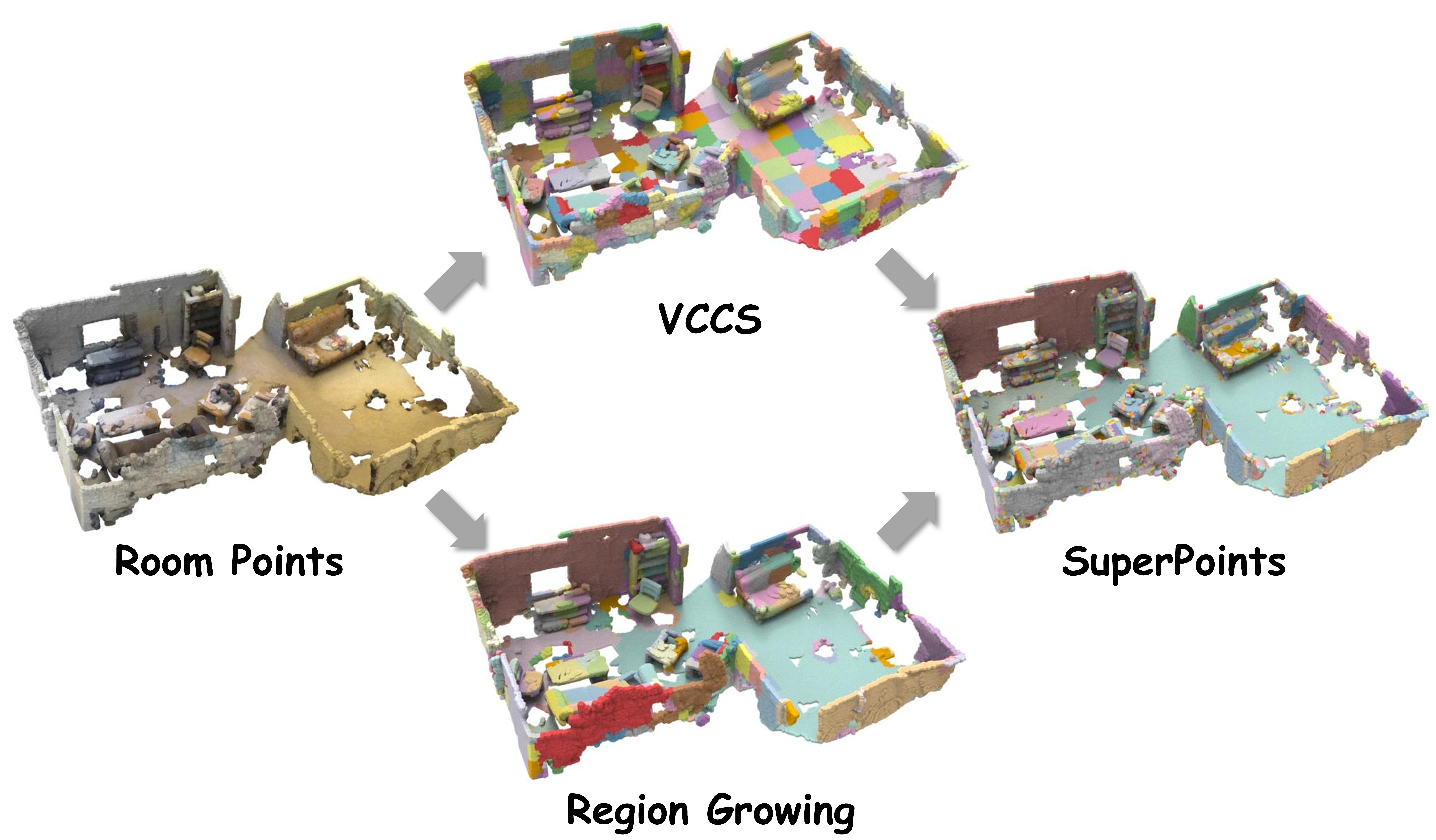}
    \vspace{-0.5cm}
    \caption{\textbf{Superpoint construction pipeline.} Starting from a room point cloud, VCCS and region growing produce complementary oversegmentations that are merged into a set of superpoints, which serve as compact geometric units for subsequent scene graph construction.}
    \label{fig:Superpoint_Construction}
    \vspace{-0.3cm}
\end{figure}

To obtain a compact geometric representation suitable for downstream graph construction and clustering, we decompose each room-level point cloud into superpoints that preserve local geometric continuity, planar smoothness, and appearance consistency. Compared with operating on raw points, superpoints substantially reduce redundancy and computation cost while serving as a stable and noise-resistant processing unit for later semantic reasoning and graph node formation.

Given a room-level point cloud ${P}_{room}$, we first normalize its coordinates by subtracting the global centroid and voxelize the 3D space to suppress noise while preserving the underlying geometry. 
As illustrated in Fig.~\ref{fig:Superpoint_Construction}, we then obtain two complementary oversegmentations from the same room points. 
The first branch applies Voxel Cloud Connectivity Segmentation (VCCS)~\cite{VCCS}, which produces fine-grained supervoxels based on a similarity measure that jointly considers spatial proximity, surface-normal consistency, and perceptual RGB distance. While VCCS effectively captures local geometric detail, it often yields fragmented segments around thin structures or depth-incomplete regions. 
In contrast, the second branch performs region growing~\cite{GrowSP} under curvature-based smoothness and neighborhood similarity constraints, generating larger and more geometry-consistent regions that better adhere to continuous surfaces. 
Leveraging the complementary strengths of these two segmentations, we adopt a consistency-based merging strategy in which each VCCS segment is reassigned to its dominant region-growing label if the latter accounts for more than half of its points; otherwise, the original VCCS label is retained. The resulting label vector $S$ defines the final set of superpoints, each representing a coherent geometric subset of ${P}_{room}$, which subsequently serve as the atomic units for fine-grained scene graph construction, semantic enrichment, and object-level clustering.

\section{Voronoi Navigation Graph Construction}
\label{sec:Voronoi_Navigation_Graph_Construction}

\begin{figure*}[t]
    \centering
    \includegraphics[width=\linewidth]{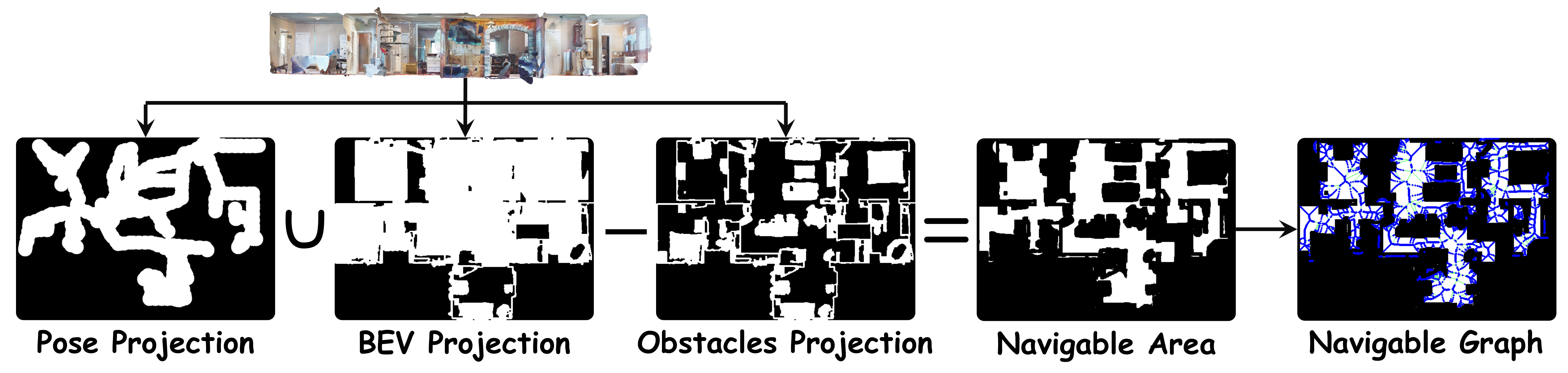}
    \vspace{-0.5cm}
    \caption{\textbf{Voronoi-based navigation graph construction.} Camera-pose and floor-support BEV projections are fused and refined by removing obstacle regions to obtain a navigable area mask. A Voronoi diagram is then computed from the free-space map, and its medial-axis skeleton is sampled to produce a sparse, topology-preserving navigation graph.}
    \label{fig:Voronoi_Navigation_Graph}
    \vspace{-0.2cm}
\end{figure*}

Before constructing the Traversable Topological Graph, we first obtain a baseline navigation topology that ensures basic route connectivity and static obstacle avoidance. To this end, we generate a Voronoi-based navigation graph~\cite{IGBATMFMRN} that captures the connectivity of free space and serves as the geometric backbone for both high-level planning and low-level execution. This foundational structure enables us to subsequently apply the Traversability Update Strategy, explicitly modeling movable obstacles and upgrading the graph from static collision-free navigation to interaction-aware traversal in movable-object environments.

As illustrated in Fig.~\ref{fig:Voronoi_Navigation_Graph}, for each floor, we estimate the navigable area on a bird’s-eye-view grid by fusing three complementary projections: camera poses, floor support, and obstacles. 
We first project all camera centers onto the horizontal plane and dilate each projection with a fixed-radius disk to obtain a pose projection map that approximates the regions actually traversed during scanning. 
In parallel, we project all floor-level points to form a BEV floor-support projection map that delineates the spatial extent of the reconstructed floor surface. Taking the union of these two maps yields a candidate floor region that is either observed by the cameras or geometrically supported. 
To account for blocking structures, we then extract 3D points lying above the floor but below a reasonable height threshold and project them to BEV to obtain an obstacle projection map. Subtracting this obstacle map from the candidate region produces the final navigable area, which is subsequently used for Voronoi-based navigation graph construction.  
From the binary navigable area mask, we first compute a 2D distance transform and generate its Voronoi diagram, whose ridges correspond to the medial axes of collision-free space. We then trace these Voronoi ridges to obtain continuous skeleton curves and sample points along them at regular spatial intervals. Connecting adjacent samples along each ridge yields a set of well-spaced waypoints that preserve the topology of the free space while avoiding redundant density. Each waypoint is then lifted back into 3D by assigning the corresponding floor height. Finally, we remove short spurious branches and isolated fragments, producing a clean, sparse, and well-connected navigation graph suitable for downstream planning.

\section{Efficiency Filtering Algorithmic Details}

\label{sec:Efficiency_Filtering_Algorithmic_Details}
\begin{figure}[t]
    \centering
    \includegraphics[width=\linewidth]{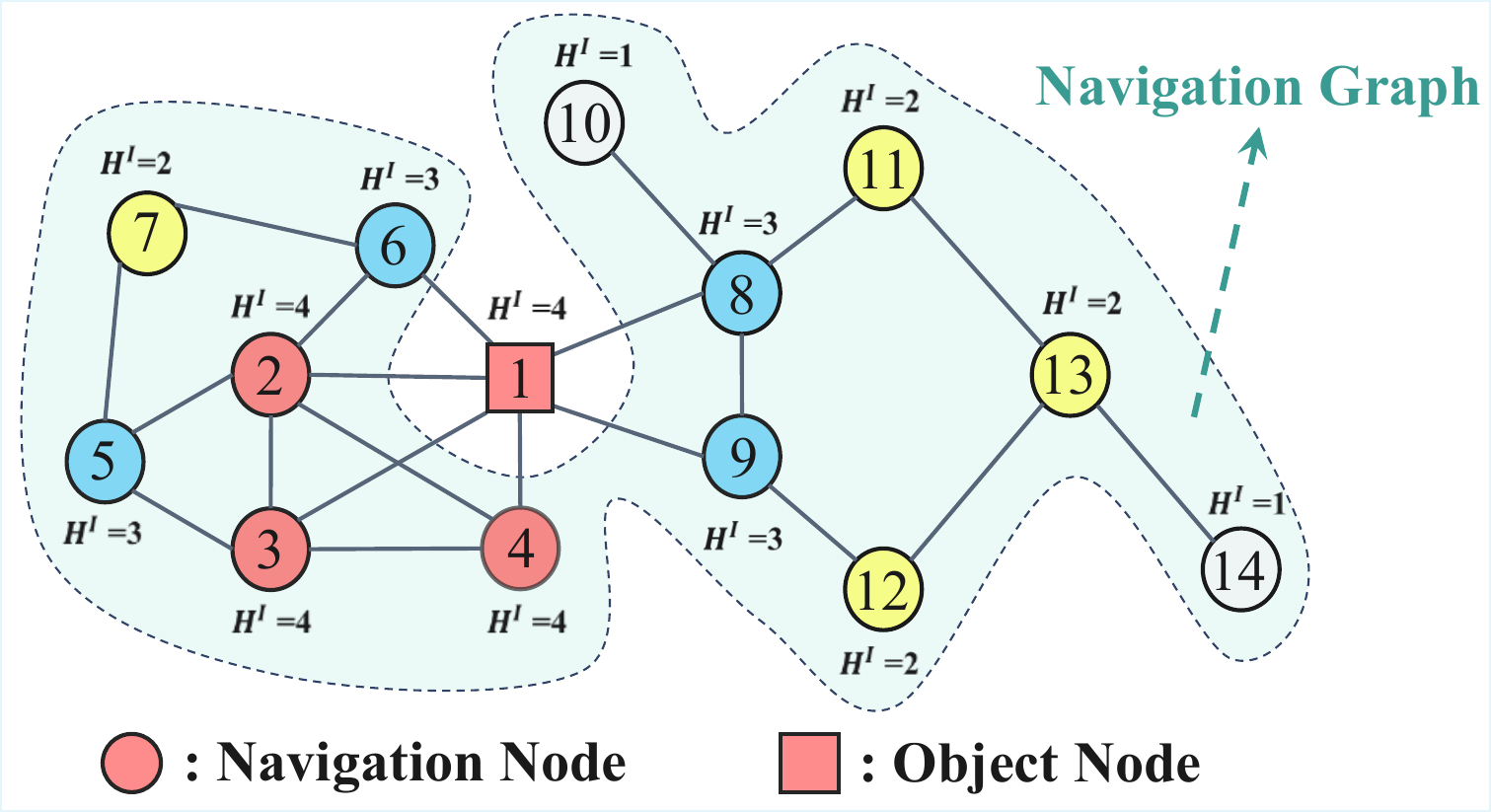}
    \vspace{-0.5cm}
    \caption{\textbf{Example of the K-Shell Iteration Factor algorithm.} K-Shell Iteration Factor algorithm is applied on the augmented navigation graph to estimate the potential navigation-efficiency gain of interacting with the candidate object node (ID~1).}
    \label{fig:appendix_KSIF}
    \vspace{-0.2cm}
\end{figure}

\subsection{Preliminaries}
The K-Shell Iteration Factor~\cite{KSIF} is built upon the classical K-Shell decomposition method. Therefore, before introducing our efficiency evaluation mechanism, we briefly revisit the fundamental concept of node degree and the hierarchical coreness analysis derived from the K-Shell decomposition.

\textbf{Degree}~\cite{degree}, denoted as $k$, is one of the earliest local metrics used for estimating node influence. It is defined by counting the number of directly connected neighboring nodes, reflecting how well a node is locally embedded within its immediate vicinity. A higher degree suggests stronger local connectivity; for example, as illustrated in Fig.~\ref{fig:appendix_KSIF}, node $1$ has a degree of $6$ ($k = 6$), because it is directly linked to six neighboring nodes.

\textbf{K-shell}~\cite{K_shell}, denoted as $k^s$, is an early used global metric for characterizing node importance from a hierarchical topological perspective. Unlike degree, which only reflects local connectivity, the K-shell decomposition method uncovers layered structural organization by iteratively peeling nodes based on their degrees. Specifically, all nodes with degree $k = 1$ are removed in the first iteration, forming the 1-shell and being assigned a coreness value of $k^s = 1$, as exemplified by nodes 10 and 11 in Fig.~\ref{fig:appendix_KSIF}.
This removal may cause remaining nodes to update their degrees and potentially drop to $k \leq 1$, in which case they are subsequently removed within the same shell. The procedure is then recursively applied to the remaining graph using degree thresholds $k = 2, 3, \dots$, thereby extracting the 2-shell, 3-shell, and higher-order shells. Through this hierarchical peeling process, each node ultimately receives a shell index $k^s$, where higher values indicate deeper embedding within the network and stronger global structural significance.

\subsection{K-Shell Iteration Factor for Efficiency Filtering}
To assess whether interacting with a candidate object can potentially improve global navigation efficiency, we adopt the K-Shell Iteration Factor as a graph-based importance evaluation metric. As illustrated in Fig.~\ref{fig:appendix_KSIF}, given a candidate object node, we temporarily insert it into the existing navigation graph, forming an augmented graph $G_t$. Our objective is to determine the relative importance ranking of the candidate object within $G_t$, such that a higher ranking implies a higher expected efficiency gain if the object is selected for interaction. 
Taking the object node 1 in the example graph as a demonstration, we follow the K-shell hierarchical peeling procedure to iteratively remove nodes, while adopting a modified value assignment scheme to compute the iterative removal depth $H^I$ for all nodes.
Specifically, all nodes with degree $k=1$ are first removed from the graph, resulting in the removal and assignment of $H^I=1$ to nodes 10 and 14. 
Next, the peeling is repeated on the remaining graph, where nodes with updated degree $k=2$ are removed, assigning $H^I=2$ to nodes 7, 11, 12, and 13. This process continues by removing nodes with degree $k=2$ in the third iteration, assigning $H^I=3$ to nodes 5, 6, 8, and 9. Finally, the remaining core nodes 1, 2, 3, and 4 are assigned $H^I=4$, indicating that they form the innermost and most structurally persistent region of the graph. Similarly, the corresponding coreness value $k^s$ for each node is also obtained during this hierarchical peeling process (e.g., the coreness of node~1 is $k^s = 3$ in this example). 

After obtaining the coreness and iteration assignments $(k^s, H^I)$ for all nodes, we then compute the K-Shell Iteration Factor for each node and rank all nodes in $G_t$ according to their $KS^{IF}$ values. The resulting ranking position of the candidate object node is interpreted as an efficiency-oriented importance score: nodes that appear closer to the top of this ranking are regarded as providing higher potential benefit to global navigation if interacted with. In our efficiency filtering module, only candidates whose KSIF-based importance exceeds a predefined threshold are considered worthwhile to interact with and are thus promoted to movable obstacles, while those with low KSIF ranks are treated as non-interactive and remain part of the static environment.

\section{Semantic Filtering Details}
\label{sec:Semantic_Filtering_Details}
\begin{figure}[t]
    \centering
    \includegraphics[width=\linewidth]{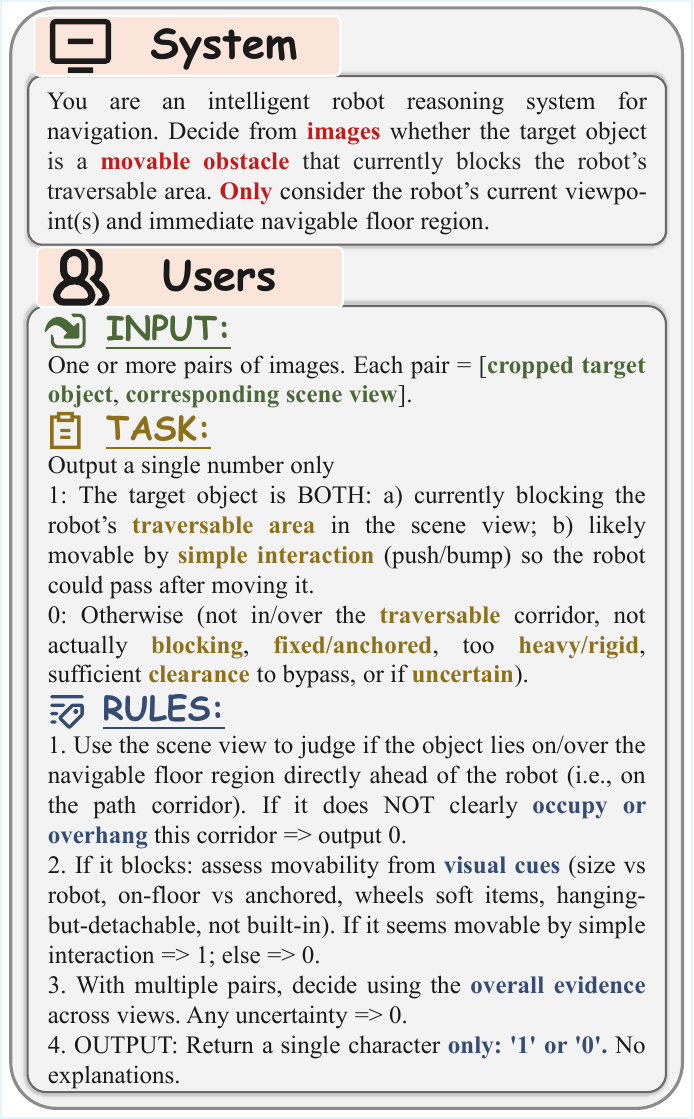}
    \vspace{-0.5cm}
    \caption{
        \textbf{Prompt design for the semantic filtering module.} The Gemini-based vision–language verifier receives paired visual inputs (cropped object and corresponding scene view) and returns a deterministic binary decision indicating whether the object should be treated as a movable obstacle..
    }
    \label{fig:Semantic_Filtering_Prompt}
    \vspace{-0.2cm}
\end{figure}

To further ensure the correctness of functional movability recognition, we employ a vision–language–guided semantic verification module to refine the candidates that pass the efficiency-driven filtering stage. This semantic filtering aims to exclude objects that, although theoretically beneficial from a topological efficiency perspective, are not physically movable in real-world conditions due to being rigid, anchored, built-in, or structurally non-operable. Specifically, we use the Gemini-2.5-Pro~\cite{gemini} vision–language model with the sampling temperature fixed at 0 to enforce deterministic binary outputs. Given one or more paired visual inputs (a cropped target-object image along with its corresponding egocentric scene view), the model is instructed to return a single binary label, 1 or 0, indicating whether the object should be regarded as a movable obstacle. Only predictions of 1 are accepted as semantically validated movable obstacles, while all others are conservatively discarded. The detailed prompt design is illustrated in Fig~\ref{fig:Semantic_Filtering_Prompt}.

\label{sec:Efficiency_Filtering_Algorithmic_Details}

\section{Details of Scene Representation Evaluation}
\label{appendix:Details_of_Scene_Representation_Evaluation}
\subsection{Implementation Details}
We assign semantic and instance labels to ground-truth (GT) points by performing a $k$-nearest neighbor search ($k=5$) in the predicted point cloud and determining the final label via majority voting. During evaluation, we exclude three types of regions: unlabeled points, wall and floor-mat. In the Topological Clustering Strategy, object-level aggregation adopts a progressive-growing merging schedule, where the similarity threshold is linearly relaxed from 0.9 to 0.5 across five stages to ensure stable small-to-large region consolidation. For feature encoding, each object node is represented using 10 multi-view image sampled from distinct viewpoints. Global semantic compensation employs a five-step scale expansion with a multiplicative growth ratio of 0.1, while local refinement selects five SAM-prompt points per object to recover missing semantics and enhance fine-grained consistency.

\subsection{Evaluation Metrics}
For evaluating the structural and semantic quality of the constructed scene graph, we adopt four standard metrics: mean Intersection-over-Union (mIoU), Frequency-weighted mean Intersection-over-Union (F-mIoU), mean Class Accuracy (mAcc), and mean Average Precision (mAP). Specifically, mIoU measures the average overlap between predicted and ground-truth semantic regions across all classes, providing a balanced assessment of segmentation quality. F-mIoU further incorporates per-class frequency to weight contributions by their occurrence, thus mitigating the influence of rare classes and better reflecting real-world scene distributions. mAcc computes the average per-class classification accuracy and reflects the model’s ability to correctly assign semantic labels irrespective of class imbalance. mAP, used for instance-level evaluation, measures the average detection precision across IoU thresholds, emphasizing object completeness and discriminability in the generated instance representations.

\section{Details of 3D Visual Grounding Evaluation}
\label{appendix:Details_of_3D_Visual_Grounding_Evaluation}
\subsection{Implementation and Variants}

\begin{figure}[t]
    \centering
    \includegraphics[width=\linewidth]{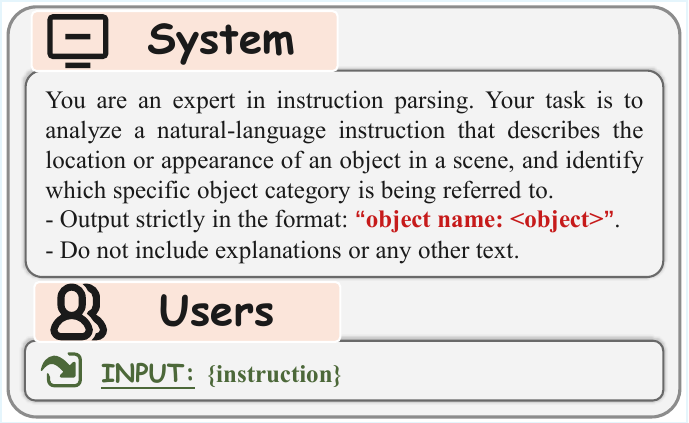}
    \vspace{-0.5cm}
    \caption{\textbf{Prompt design for referential object extraction.} GPT-5 is prompted to identify the object category referenced in a full natural-language instruction.}
    \label{fig:prompt_gpt_5}
    \vspace{-0.1cm}
\end{figure}

We retrieve the target by computing the cosine similarity between the CLIP-encoded~\cite{chen2024clip} full natural-language instruction and the semantic embeddings of the three object-node candidates, without any simplification or preprocessing. In addition to the Full Instruction retrieval setting, we further compare two Part Instruction variants as show in Table~\ref{tab:appendix_3dvg}. 
In the first Part Instruction (CLIP) setting, we encode only the explicit object category mentioned in the instruction using CLIP and conduct similarity-based retrieval. 
In the second Part Instruction (CLIP + GPT-5) setting, we first employ a GPT-5 based linguistic extractor to infer the referential object type from the full description, using the prompt design illustrated in Fig.~\ref{fig:prompt_gpt_5}, and then encode the extracted noun phrase using CLIP for similarity matching.
Experimental results indicate that directly encoding the complete natural-language instruction yields the highest retrieval accuracy. This indicates that the semantics encoded by our framework preserve globally coherent feature representations and remain robust to redundant or over-complete descriptions, enabling more accurate retrieval by exploiting the contextual semantics conveyed by full-sentence instructions.

\subsection{Evaluation Metrics}
We adopt two widely used metrics, Acc@0.25 and Acc@0.5, to measure the correctness of object localization with respect to language queries. Specifically, Acc@0.25 denotes the percentage of grounding results whose predicted 3D bounding box achieves an Intersection-over-Union (IoU) with the ground-truth box greater than 0.25, providing a relatively tolerant assessment that reflects coarse yet semantically aligned localization capability. In contrast, Acc@0.5 tightens the IoU threshold to 0.5, quantifying fine-grained and spatially precise localization performance.

\section{Details of Interactive Navigation Evaluation}
\label{appendix:Details_of_Interactive_Navigation_Evaluation}
\begin{table}[t]
    \centering
    \caption{ \textbf{Evaluation of 3D visual grounding variants.} We compare full-instruction encoding with two part-instruction baselines, where only the explicit object category (CLIP) or a GPT-5 extracted noun phrase (CLIP+GPT-5) is encoded.}
    \vspace{-0.2cm}
    \resizebox{\linewidth}{!}{
    \begin{tabular}{c|c|cc}
        \toprule
        \textbf{~~Description~~}  & \textbf{~~Agent~~}  & \textbf{~~Acc@$\mathbf{0.25}$~~} & \textbf{~~Acc@$\mathbf{0.5}$~~} \\
        \midrule\midrule
        \multirow{2}{*}{Part Instruction} 
            & CLIP + GPT-5 & 49.7 & 36.2 \\
            & CLIP & \underline{56.1} & \underline{41.5} \\
        \midrule
        Full Instruction  & CLIP & \textbf{58.3} & \textbf{43.7} \\
        \bottomrule
    \end{tabular}
    }
    \vspace{-0.2cm}
    \label{tab:appendix_3dvg}
\end{table}

\begin{table}[t]
    \centering
    \caption{\textbf{Details of the augmented scenes.} This table provides the correspondence between each augmented scene and its original HM3D identifier, as well as the types of movable obstacles introduced in each environment.}
    \vspace{-0.2cm}
    \resizebox{\linewidth}{!}{
    \begin{tabular}{c|c|c}
        \toprule
        \textbf{Scene ID} & \textbf{HM3D ID} & \textbf{movable obstacles} \\
        \midrule\midrule
        1 & 00856-FnSn2KSrALj & Carton\_1, Trolley\_1, Trolley\_4, Ball\_1\\
        2 & 00824-Dd4bFSTQ8gi & Carton\_1, Screen\_1\\
        3 & 00894-HY1NcmCgn3n & Carton\_3, Carton\_4, Screen\_1, Trolley\_2\\
        4 & 00827-BAbdmeyTvMZ & Carton\_5, Carton\_6, Screen\_3\\
        5 & 00848-ziup5kvtCCR & Carton\_7, Screen\_4\\
        6 & 00829-QaLdnwvtxbs & Screen\_2\\
        7 & 00880-Nfvxx8J5NCo & Carton\_2, Screen\_5\\
        8 & 00883-u8ug2rtNARf & Screen\_2, Ball\_2\\
        
        \bottomrule
    \end{tabular}}
    \vspace{-0.2cm}
    \label{tab:appendix_scene}
\end{table}

\begin{figure*}[t]
    \centering
    \includegraphics[width=\linewidth]{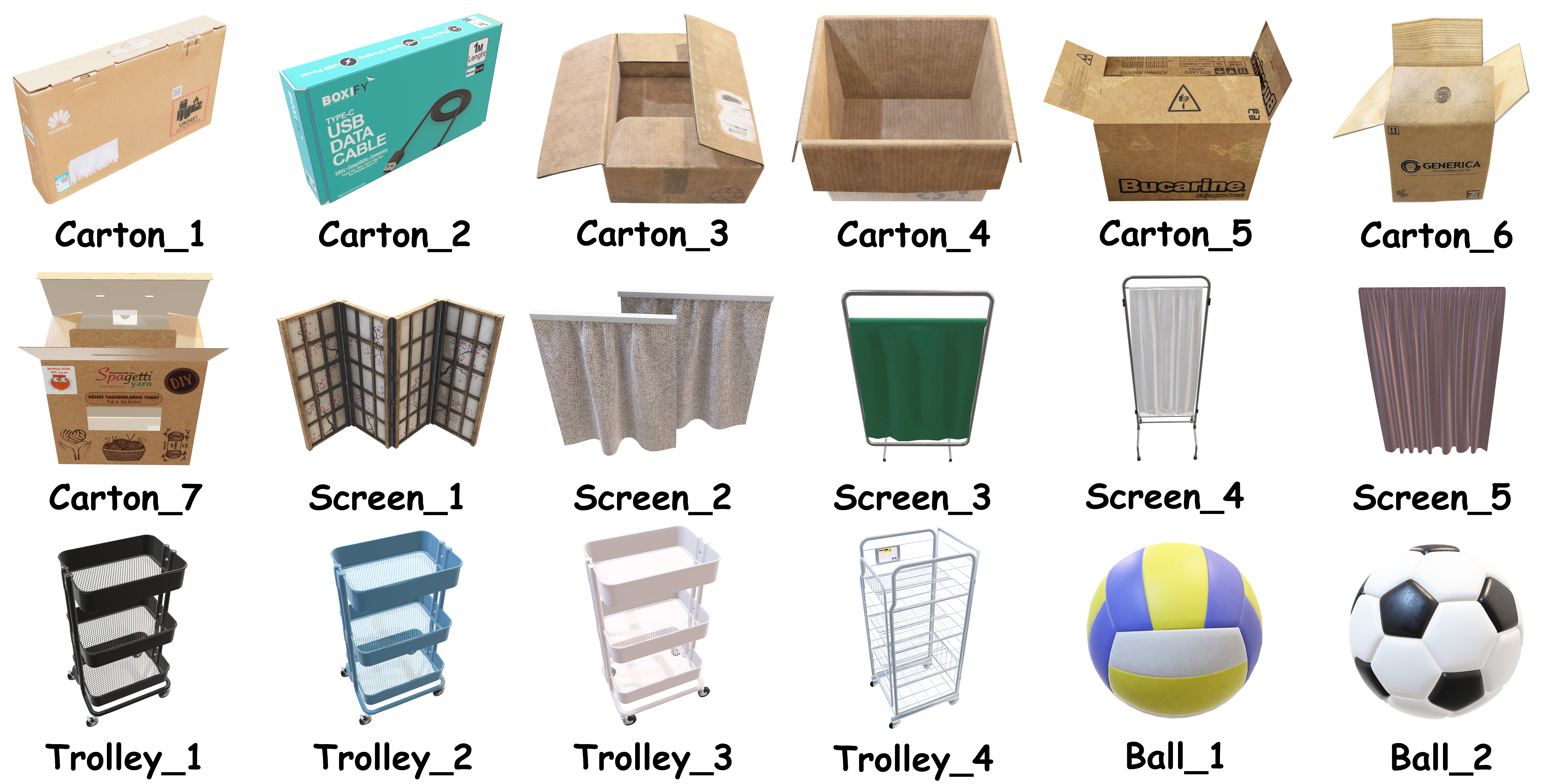}
    \vspace{-0.5cm}
    \caption{\textbf{Movable obstacles used in the augmented scenes.}
    The benchmark includes 18 manually curated movable-obstacle instances spanning four representative categories: carton, screen, trolley, and ball, which are inserted into key traversable regions of the HM3D environments to create realistic interaction-driven navigation scenarios.}
    \label{fig:Scene_obstacles}
    \vspace{-0.1cm}
\end{figure*}

\begin{table*}[t]
    \centering
    \caption{\textbf{Interactive navigation task set.} This table summarizes the human-written referring instructions used to construct interactive navigation tasks across the augmented scenes. In the instruction text, \textbf{\textcolor{teal!60!cyan!60!white}{teal}} highlights object descriptions and \textbf{\textcolor{orange!50}{orange}} marks room-related cues. For each scene, both concise referring expressions and more detailed descriptive instructions are provided, and each instruction is further instantiated into multiple tasks by assigning different starting positions.}
    \vspace{-0.2cm}
    \resizebox{\linewidth}{!}{
    \begin{tabular}{c|l}
        \toprule
        \textbf{ID}  & \multicolumn{1}{c}{\textbf{Instruction}}\\
        \midrule\midrule

        \multirow{5}{*}{1}  & Find the \textbf{\textcolor{teal!60!cyan!60!white}{mirror}} in the \textbf{\textcolor{orange!50}{bathroom}} \\
        
        & Find the \textbf{\textcolor{teal!60!cyan!60!white}{garbage bin}} in the \textbf{\textcolor{orange!50}{living room}} \\
        
        & Find the \textbf{\textcolor{teal!60!cyan!60!white}{dinning table}} in the \textbf{\textcolor{orange!50}{living room}} \\
        
        & Find a \textbf{\textcolor{teal!60!cyan!60!white}{brown basket}} in a \textbf{\textcolor{orange!50}{room with a blackboard}} \\
        
        & Find a \textbf{\textcolor{teal!60!cyan!60!white}{small wooden stool next to the sofa}} in a \textbf{\textcolor{orange!50}{living room with an open kitchen}} \\
        \midrule

        \multirow{5}{*}{2}  & Find the \textbf{\textcolor{teal!60!cyan!60!white}{bed}} in the \textbf{\textcolor{orange!50}{bedroom}} \\

        & Find the \textbf{\textcolor{teal!60!cyan!60!white}{sofa}} in the \textbf{\textcolor{orange!50}{living room}} \\

        & Find the \textbf{\textcolor{teal!60!cyan!60!white}{globe}} in the \textbf{\textcolor{orange!50}{study room}} \\
        
        & Find the \textbf{\textcolor{teal!60!cyan!60!white}{dining-table}} in the \textbf{\textcolor{orange!50}{restaurant}} \\

        & Find the \textbf{\textcolor{teal!60!cyan!60!white}{flower pot on the coffee table}} in the \textbf{\textcolor{orange!50}{room with an open-plan kitchen and living room}} \\
        \midrule

        \multirow{5}{*}{3}  & Find a \textbf{\textcolor{teal!60!cyan!60!white}{gold and black shield}} in the \textbf{\textcolor{orange!50}{main exhibition hall}} \\
        
        & Find the \textbf{\textcolor{teal!60!cyan!60!white}{golden female statue}} in the \textbf{\textcolor{orange!50}{room with black curtains}} \\
        
        & Find a \textbf{\textcolor{teal!60!cyan!60!white}{brown wooden carved crucifix}} in the \textbf{\textcolor{orange!50}{room with two cardboard boxes}} \\
        
        & Find the \textbf{\textcolor{teal!60!cyan!60!white}{middle chair among three chairs placed side by side}} in the \textbf{\textcolor{orange!50}{room with a brown crucifix}} \\
        
        & Find a \textbf{\textcolor{teal!60!cyan!60!white}{solemn Virgin Mary wearing a black and gold robe, holding a sword, with a radiant halo and an ornate altar background}} in the \textbf{\textcolor{orange!50}{room with black curtains}} \\
        \midrule

        \multirow{5}{*}{4}  & Find the \textbf{\textcolor{teal!60!cyan!60!white}{sofa}} in the \textbf{\textcolor{orange!50}{lounge}}. \\

        & Find the \textbf{\textcolor{teal!60!cyan!60!white}{refrigerator}} in the \textbf{\textcolor{orange!50}{main hall}}. \\
        
        & Find the \textbf{\textcolor{teal!60!cyan!60!white}{chair}} in the \textbf{\textcolor{orange!50}{secondary bedroom}}. \\
        
        & Find the \textbf{\textcolor{teal!60!cyan!60!white}{bowl on the dining table}} in the \textbf{\textcolor{orange!50}{main hall}}. \\

        & Find the \textbf{\textcolor{teal!60!cyan!60!white}{laundry detergent on the washing machine}} in the \textbf{\textcolor{orange!50}{main hall}}. \\
        \midrule

        \multirow{5}{*}{5}  & Find the \textbf{\textcolor{teal!60!cyan!60!white}{toilet}} in the \textbf{\textcolor{orange!50}{en-suite bathroom}}. \\
        
        & Find the \textbf{\textcolor{teal!60!cyan!60!white}{kitchen sink}} in the \textbf{\textcolor{orange!50}{open-plan kitchen}}. \\
        
        & Find the \textbf{\textcolor{teal!60!cyan!60!white}{armchair}} in the \textbf{\textcolor{orange!50}{master bedroom with a TV}}. \\

        & Find the \textbf{\textcolor{teal!60!cyan!60!white}{cabinet}} in the \textbf{\textcolor{orange!50}{utility room with a washing machine}}. \\
        
        & Find the \textbf{\textcolor{teal!60!cyan!60!white}{bed}} in the \textbf{\textcolor{orange!50}{guest bedroom with a full-length mirror, a TV, and a desk}}. \\

        \midrule

        \multirow{3}{*}{6} & Find the \textbf{\textcolor{teal!60!cyan!60!white}{TV}} in the \textbf{\textcolor{orange!50}{living room}}. \\
        
        & Find the \textbf{\textcolor{teal!60!cyan!60!white}{white pajamas}} in the \textbf{\textcolor{orange!50}{walk-in closet}}. \\
        
        & Find the \textbf{\textcolor{teal!60!cyan!60!white}{bathtub}} in the \textbf{\textcolor{orange!50}{ensuite bathroom with a shower in the master bedroom}}. \\
        \midrule

        \multirow{5}{*}{7}  & Find the \textbf{\textcolor{teal!60!cyan!60!white}{TV}} in the \textbf{\textcolor{orange!50}{master room}}. \\
        
        & Find the \textbf{\textcolor{teal!60!cyan!60!white}{trash can}} in the \textbf{\textcolor{orange!50}{living room}}. \\

        & Find the \textbf{\textcolor{teal!60!cyan!60!white}{dog bowl}} in the \textbf{\textcolor{orange!50}{family room}}. \\

        & Find the \textbf{\textcolor{teal!60!cyan!60!white}{table football}} in the \textbf{\textcolor{orange!50}{recreation room}}. \\

        & Find the \textbf{\textcolor{teal!60!cyan!60!white}{clothes hanger}} in a \textbf{\textcolor{orange!50}{walk-in closet filled with clothes}}. \\
        \midrule

        \multirow{5}{*}{8} & Find the \textbf{\textcolor{teal!60!cyan!60!white}{brown kitchen sink}} in the \textbf{\textcolor{orange!50}{kitchen}}. \\

        & Find the \textbf{\textcolor{teal!60!cyan!60!white}{washing machine}} in a \textbf{\textcolor{orange!50}{utility room}}. \\
        
        & Find the \textbf{\textcolor{teal!60!cyan!60!white}{doll}} in a \textbf{\textcolor{orange!50}{lounge with a light green sofa}}. \\

        & Find a \textbf{\textcolor{teal!60!cyan!60!white}{doll sitting on a pink sofa}} in a \textbf{\textcolor{orange!50}{bedroom with red fabric on the bed}}. \\

        & Find an \textbf{\textcolor{teal!60!cyan!60!white}{antique display cabinet filled with valuable artworks}} in the \textbf{\textcolor{orange!50}{living room}}. \\

        \bottomrule
    \end{tabular}}
    \vspace{-0.1cm}
    \label{tab:appendix_task}
\end{table*}

\subsection{Scene Definition}
We construct an augmented benchmark based on eight indoor scenes from the HM3D~\cite{HM3D} dataset.
For each selected scene, we introduce a set of visually identifiable and physically operable objects into key traversable regions, forming realistic movable obstacles that may require interaction-driven decision making rather than purely collision-free planning. 
As illustrated in Fig.~\ref{fig:Scene_obstacles}, the inserted items consist of four representative categories: carton, screen, trolley, and ball, resulting in a total of 18 movable-obstacle instances.
Across the eight scenes, between two and five movable obstacles are placed per environment, and each obstacle is manually positioned at critical spatial chokepoints where failing to interact could lead to substantial detours or even render the target region unreachable. 
As shown in Table~\ref{tab:appendix_scene}, we also provide the mapping between each augmented scene and its original HM3D identifier along with the corresponding movable-obstacle instance IDs. 

\subsection{Task Definition}
Across the augmented scenes, we define multiple interactive navigation tasks for each environment. As summarized in Table~\ref{tab:appendix_task}, the agent receives a human-written referring instruction and must navigate to the corresponding target location. For every scene, we include both concise referring expressions and more detailed descriptive instructions, where each instruction is instantiated into 4–5 tasks by assigning different starting positions. All tasks are carefully designed so that, in most cases, passing through one or more movable obstacles is required to obtain a shorter or even feasible route, enabling a clear evaluation of navigation efficiency and reachability under movable-obstacle conditions.

\subsection{Baseline Setting}

We adopt a modified HOV-SG~\cite{HOVSG} pipeline as the representative non-interactive navigation baseline, where floor–room decomposition and room-level semantic encoding strictly follow the original formulation (without our Visibility Purification Strategy). For object-node construction, we employ our Topological Clustering Strategy to maintain consistent fine-grained scene representation, ensuring fair comparability with our method and preventing failures caused by inconsistent or fragmented semantics, so that performance differences can be attributed purely to interaction-level decision making rather than scene representation errors. For navigation construction, the baseline relies solely on the standard Voronoi-based graph without our Traversability Update Strategy.

\subsection{Implementation Details}
We perform all interactive navigation experiments in Habitat-Sim. To enable construction of the Hierarchical Traversable 3D Scene Graphs, we first collect RGB-D observations and corresponding camera poses using a virtual sensing setup equipped with an onboard RGB-D camera (1080×720 resolution, 1.5 m height, 90° HFOV). To ensure sufficient multi-view coverage for geometric reasoning and semantic aggregation, the agent acquires panoramic observations by moving 0.2 m per step and rotating 5° per turn. During evaluation, an episode is considered successful if the agent terminates within 1.5 m Euclidean distance of the target location.

\subsection{Evaluation Metrics}
We adopt four metrics to evaluate interactive navigation performance: Path Length (PL), Navigation Error (NE), Success weighted by Path Length (SPL), and Success Rate (SR). Among them, PL serves as the primary indicator of navigation efficiency and is computed by first identifying the intersection of successful task sets from both the baseline and our method, and then averaging the executed trajectory lengths within this shared subset; this design ensures a fair comparison by eliminating bias introduced by uneven task success and highlights whether interaction-aware planning can genuinely shorten traversal. NE reports the terminal Euclidean distance between the agent and the target across all trials. SPL jointly considers success and path optimality by rewarding short successful trajectories while penalizing detours. SR simply measures the percentage of successful episodes and reflects global reachability, especially under blocked or partially blocked conditions.

\section{Ablation Study}
\label{appendix: Ablation_Study}

\subsection{Details of the Topological Clustering Strategy }
To evaluate the impact of the Topological Clustering Strategy on the overall system performance, we construct controlled variants by selectively disabling its internal modules and replacing them with simplified counterparts. 
When the object-node construction module is removed, we replace it with a purely 2D-driven baseline, where instance masks are extracted from RGB frames using SAM~\cite{sam}, directly projected into 3D, and the resulting raw point-cloud fragments are treated as object nodes without any topological merging or geometric consistency enforcement. Similarly, when the object-node encoding module is disabled, we replace the semantic enhancement pipeline with a direct multi-view embedding baseline, where all viewpoints observing a given object node are encoded using CLIP, and the resulting features are aggregated via averaging to form a single semantic representation. 
In addition, we assess encoder compatibility by comparing two representative models from the \url{https://github.com/mlfoundations/open_clip}, namely CLIP (ViT-H-14) and SigLIP~\cite{siglip} (ViT-SO400M-14-SigLIP), using their official pretrained checkpoints.

\subsection{Effect of the Visibility Purification Strategy}
\begin{table}[t]
    \centering
    \caption{\textbf{Ablation of the Visibility Purification Strategy.} Room-retrieval success rates comparing HERO with and without Visibility Purification Strategy (VPS) on Simple, Complex, and overall query sets.}
    \vspace{-0.2cm}
    \resizebox{\linewidth}{!}{
    \begin{tabular}{c|ccc}
        \toprule
        \textbf{~~~~Method~~~~}  & \textbf{~~~Simple~~~}  & \textbf{~~~Complex~~~} & \textbf{~~~All~~~} \\
        \midrule\midrule
        HERO(w/o VPS)  & 8/10 & 6/10 & 14/20 \\
        HERO(w/ VPS)  & 10/10 & 10/10 & 20/20 \\
        \bottomrule
    \end{tabular}
    }
    \vspace{-0.2cm}
    \label{tab:vps_ablation}
\end{table}

To evaluate the effectiveness of the proposed Visibility Purification Strategy for room-level representation, we design a room retrieval task on two representative scenes (ID~2 and ID~8). 
For each scene, the agent receives a natural-language description and must retrieve the corresponding room node. The queries are divided into two categories: Simple and Complex. 
Simple queries use coarse room-type descriptions such as “bedroom” or “kitchen” where multiple rooms in the scene may satisfy the category and retrieving any valid match is counted as success. 
Complex queries, in contrast, specify a unique target room by adding fine-grained appearance or layout cues, for example, “A room featuring a floral carpet and a chair placed on it.”

As shown in Table~\ref{tab:vps_ablation}, the Visibility Purification Strategy yields consistent and notable improvements across both query types. The success rate increases by 20\% for Simple queries and 40\% for Complex queries, leading to a 30\% overall improvement. These results demonstrate that the Visibility Purification Strategy not only mitigates semantic drift by suppressing cross-room contamination, but also strengthens the discriminative capability of room representations, enabling them to better preserve room-specific semantic characteristics.

\subsection{Effect of the Traversability Update Strategy}



        

\begin{table}[t]
    \centering
    \caption{\textbf{Ablation of the Traversability Update Strategy.} We evaluate how Efficiency Filtering (EF), Semantic Filtering (SF), and different vision–language model backends affect recognition accuracy (RA) and the number of model invocations (\#Calls) for movable-obstacle identification.}
    \vspace{-0.2cm}
    \resizebox{\linewidth}{!}{
    \begin{tabular}{c|cc|cc|cc}
        \toprule
        \multirow{2}{*}{\textbf{Type}} & \multicolumn{2}{c|}{\textbf{Modules}} & \multicolumn{2}{c|}{\textbf{VLM}} & \multicolumn{2}{c}{\textbf{Statistics}} \\
        \cmidrule(lr){2-7}
        & \textbf{EF} & \textbf{SF} & \textbf{Gemini} & \textbf{GPT-4o} & \textbf{\#Calls $\downarrow$} & \textbf{RA $\uparrow$ (\%)}\\
        \midrule\midrule

        \multirow{4}{*}{\rotatebox{90}{\small Variants}} 
        & \ding{55} & \ding{55} & -- & -- & -- & 3.92 \\
        & \ding{51} & \ding{55} & -- & -- & -- & 14.71 \\
        & \ding{55} & \ding{51} & \ding{55} & \ding{51} & 121 & 10.81 \\
        & \ding{55} & \ding{51} & \ding{51} & \ding{55} & 121 & 17.24 \\

        \midrule

        \multirow{2}{*}{\rotatebox{90}{\small Ours}} 
        & \ding{51} & \ding{51} & \ding{55} & \ding{51} & 34 & 33.33 \\
        & \ding{51} & \ding{51} & \ding{51} & \ding{55} & 34 & 41.67\\
        
        \bottomrule
    \end{tabular}}
    \vspace{-0.2cm}
    \label{tab:tus}
\end{table}

To assess how the components of the Traversability Update Strategy influence the reliable and efficient identification of movable obstacles and how these choices affect downstream interactive navigation, we conduct a controlled ablation on a representative environment (Scene ID 3). 
We report two metrics: recognition accuracy (RA), defined as the proportion of predicted movable obstacles that are truly movable. This metric reflects how reliably the method selects operable objects for interaction; and the number of VLM invocations (\#Calls), which captures the computational cost of semantic verification.
We evaluate several variants by selectively enabling or disabling Efficiency Filtering (EF) and Semantic Filtering (SF). A degenerate baseline disables both modules and randomly assigns movability labels. Additional variants activate only EF or only SF, and for the SF-only setting we compare two VLM backbones (GPT-4o~\cite{gpt4o} and Gemini~\cite{gemini}) to examine their effect on accuracy and cost. These configurations together clarify how each component contributes to the reliability and efficiency of movable-obstacle identification.

As shown in Table~\ref{tab:tus}, The ablation results show that Efficiency Filtering and Semantic Filtering are strongly complementary and jointly crucial for robust movable-obstacle identification. The full strategy achieves much higher RA than using either module alone, improving over the EF-only variant by 26.96 \% and over the SF-only variant (under the same VLM configuration) by 24.43\%, while requiring only 34 VLM calls, which is about 3.5 times fewer than the SF-only settings. This indicates that EF effectively removes low-value candidates early, reducing semantic-verification cost without compromising precision. The comparison between the two VLM backends within our full configuration further shows that Gemini integrates more effectively with the Semantic Filtering module, yielding an 8.34\% higher RA than GPT-4o under identical settings.

\end{document}